\documentclass[10pt,journal,compsoc]{IEEEtran}

\usepackage{graphicx}
\usepackage{amsmath,amssymb}
\usepackage{graphicx,subfigure}
\usepackage{algorithm}
\usepackage{algorithmic}
\usepackage{hyperref}

\makeatletter

\newcommand{\Rmnum}[1]{\expandafter\@slowromancap\romannumeral #1@}
\makeatother

\newcommand{\ie}{{\it i.e.}}
\newcommand{\eg}{{\it e.g.}}

\begin{document}

\let\rm=\rmfamily    \let\sf=\sffamily    \let\tt=\ttfamily
\let\it=\itshape     \let\sl=\slshape     \let\sc=\scshape
\let\bf=\bfseries

%\title{Randomized Locality Sensitive Vocabularies for Bag-of-Features Model} % Replace with your title
%\title{Randomized Visual Vocabulary Construction with Hashing}
\title{Video De-fencing}

\author{Yadong~Mu, ~\IEEEmembership{Member,~IEEE}, Wei Liu and Shuicheng~Yan,~\IEEEmembership{Member,~IEEE}
\IEEEcompsocitemizethanks{
\IEEEcompsocthanksitem Yadong Mu is with the Department
of Electrical Engineering, Columbia University, New York, 10027, USA. Part of the work was done when the first author worked at Peking University and National University of Singapore. \protect\\
E-mail: ym2372@columbia.edu.
\IEEEcompsocthanksitem Wei Liu is currently Josef Raviv Memorial Postdoctoral Fellow of IBM T. J. Watson Research Center. \protect\\
E-mail: wliu@ee.columbia.edu.
\IEEEcompsocthanksitem Shuicheng Yan is with the Department
of Electrical and Computer Engineering, National University of
Singapore, 117576, Singapore.\protect\\
E-mail: eleyans@nus.edu.sg.
}
}

%\thanks{\textbf{Acknowledgment}: This work is partially supported by project grant NRF2007IDM-IDM002-069 on
%``Life Spaces" from the IDM Project Office, Media Development Authority of Singapore and also NRF/IDM Program under research Grant NRF2008IDMIDM004-029.}
%}

\IEEEcompsoctitleabstractindextext{
\begin{abstract}
This paper describes and provides an initial solution to a novel video editing task, \ie, \emph{video de-fencing}. It targets automatic restoration of the video clips that are corrupted by fence-like occlusions during capture. Our key observation lies in the visual parallax between fences and background scenes, which is caused by the fact that the former are typically closer to the camera. Unlike in traditional image inpainting, fence-occluded pixels in the videos tend to appear later in the temporal dimension and are therefore recoverable via optimized pixel selection from relevant frames. To eventually produce fence-free videos, major challenges include cross-frame sub-pixel image alignment under diverse scene depth, and ``correct" pixel selection that is robust to dominating fence pixels. Several novel tools are developed in this paper, including soft fence detection, weighted truncated optical flow method and robust temporal median filter. The proposed algorithm is validated on several real-world video clips with fences.
\end{abstract}

}

\maketitle
\IEEEdisplaynotcompsoctitleabstractindextext
\IEEEpeerreviewmaketitle

\section{Introduction}

\begin{figure*}[t!]
\centering
  \includegraphics[width=0.9\linewidth]{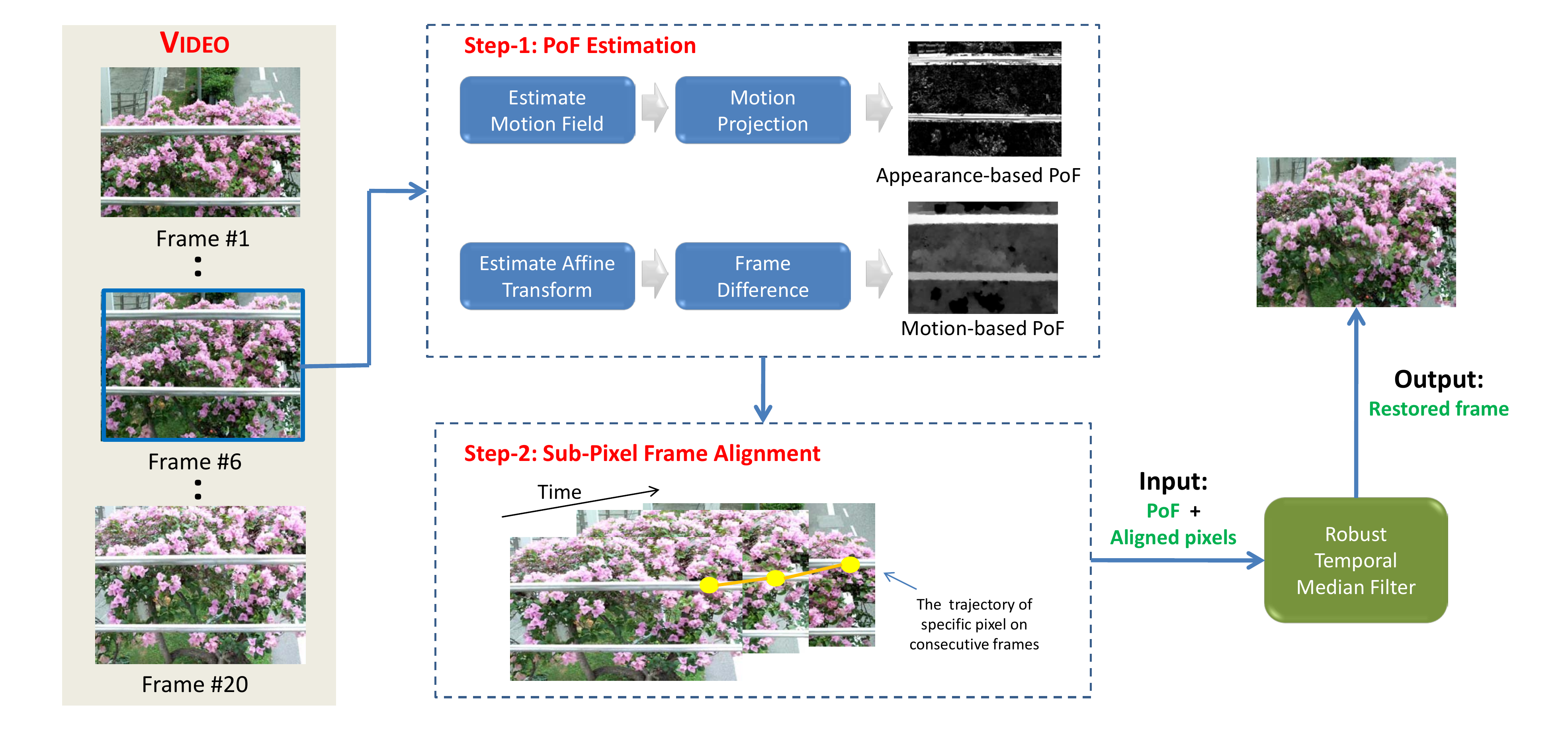}
\caption{\footnotesize Illustration of the algorithmic pipeline for video de-fencing. It shows the operations on a frame selected from fence-corrupted video clips ``Flower". All the operations (including fence detection and removal) are fully automatic.}
\label{fig:pipeline}
\end{figure*}

This paper describes a novel algorithm for automatic fence detection and removal in consumer video clips. We term this task to be ``video de-fencing". This technique can be especially useful in a variety of scenarios. When the target scenes that the users plan to capture are occluded by a fence and the users are not allowed to cross the fence (\eg, use a video camera to capture a tiger confined in the cage at the zoo), a natural solution is to record the scenes with fences and resort to specific post-processing algorithms for fence removal. As shown later in the paper, the implication of ``fence" can be largely generalized, which makes the proposed technique general enough for the daily use of the digital cameras. With the video de-fencing technique, users are able to obtain a fence-free video with litter additional effort. Moreover, recent trend in digital camera has shown the power of incorporating sophisticated algorithms into the camera hardware (\eg, high dynamic range photos). After the technique of video de-fencing becomes much more mature, it can be integrated as a part of the consumer camera hardware and attains real-time video re-touching.

% Fig.~\ref{fig:example} illustrates two examples.

For those scenes occluded by fences, the goal of video de-fencing is to automatically restore them and return fence-free videos. There is a vast amount of research that is devoted to pattern (\eg, near-regular structures, rains~\cite{garg04,garg05}, snowflakes~\cite{peter07}) detection and removal in both images and videos. Nonetheless, the video de-fencing problem is novel since fence-like structures have been seldom explored in video editing. The most relevant work to ours is the ``image de-fencing" by Liu et al.~\cite{liuyanxi08}, where the authors propose to solve the image de-fencing by two steps. First, the fences are detected according to spatial regularity (\eg, symmetry) and image masks are subsequently constructed. Afterwards, it utilizes sophisticated image inpainting techniques~\cite{marcelo00} for fence filling. The major limitation of the method stems from the assumption of repeating-texture of occluded scenes. The restoration of missing information within a single image is not a well-defined problem for general images, since in most cases the repeating-texture assumption fails to hold. In the follow-up work of~\cite{park10}, the authors further make attempt to overcome the afore-mentioned limitation by using multi-view images. Given an image $I$, it computes the optical flow (based on Lucas Kanade algorithm~\cite{Lucas81aniterative}) to another image from related view. The flow field is then used to aid finding patch correspondence across multi-view images. In this way the scarcity of source information for fence filling is partially mitigated. The method is possible to generate plausible results on some images. However, note that the method does not capitalize on the temporal information contained in the videos, especially not explicitly addressing the issue of visual parallax, which fundamentally differs from our proposed problem setting and corresponding solutions. Another straightforward solution for video de-fencing is to manually mask out the fences and perform video completion~\cite{shiratori06,wexler07}. However, mask generation is known to be labor-intensive, especially for those web-like, thin fences.

\begin{figure*}[thb]
\centering
\includegraphics[width=0.95\linewidth]{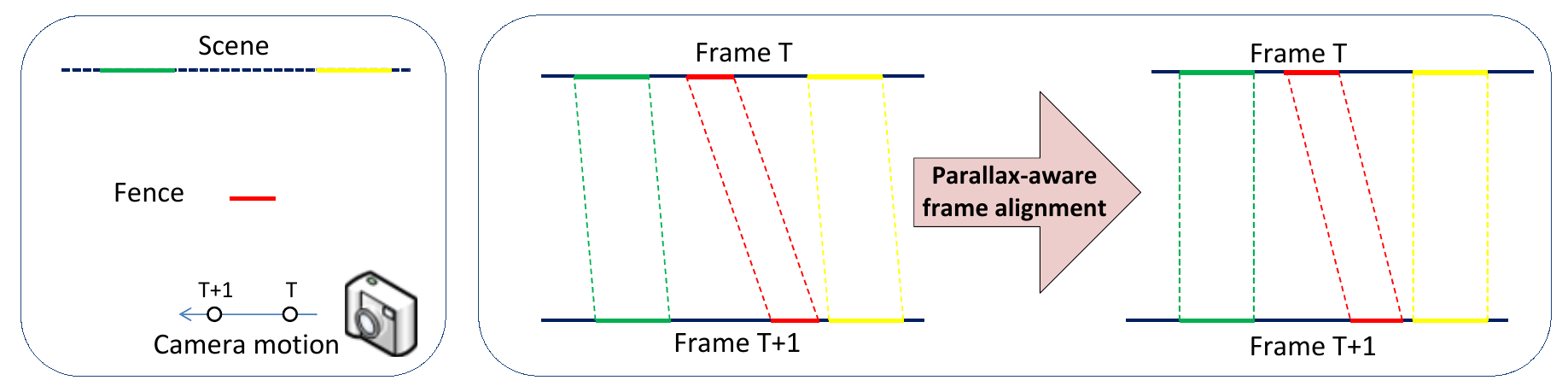}
\caption{\footnotesize Illustration of the underlying mechanism of video de-fencing. The left figure depicts the relative positions between the scene (containing two objects denoted in green and yellow respectively), fence (denoted in red) and camera, together with the camera motion path at time $T$ and $T+1$ respectively. The right figure intuitively explains the parallax phenomena caused by diverse scene depth, whose right panel further shows the situation after parallax-aware frame alignment. Namely, the frame alignment is accomplished in such a way that scene objects are well aligned across multiple frames. From the principal of parallax, the fences will fail to be aligned and it therefore enables fence detection and removal.}
\label{fig:model}
\end{figure*}

Our research is inspired by the significant advances in the field of computational photography\footnote{Visit \url{http://en.wikipedia.org/wiki/Computational\_photography} for a quick reference}. By varying specific camera parameters (\eg, flash, view points, shutter, depth of field) within a small range during image or video capture, various difficult tasks can be simplified. An example is to solve image denoising and detail transfer under low-lighting conditions by combining the strength of flash/no-flash image pairs~\cite{georg04}. For video de-fencing, the visual parallax~\cite{steinman} is observed when the depth of field is large, which provides useful cue for video analysis if the motion path of camera follows specific patterns (\eg, roughly parallel to the fences and scenes), as illustrated in Fig.~\ref{fig:model}. It is shown that \emph{fence pixels} (\ie, pixels from the image regions that correspond to the fence) show stronger drifting tendency compared with \emph{scene pixels} (\ie, pixels of the image except for fence pixels), which is known as visual parallax and serves as the basic cue to distinguish these two kinds of pixels. Moreover, under a large variety of real-world scenarios the camera moves in such a way that it is guaranteed with high probability that each pixel from scene objects is only occluded in partial of the frames and becomes visible in others, enabling occlusion removal via pixel selection from relevant frames.

Our main contribution is the exposition of an integrated video de-fencing algorithm, which consists of three successive steps: 1) estimating \emph{probability of fence} (PoF) for each pixel, 2) parallax-aware sub-pixel image alignment via the proposed \emph{weighted truncated optical flow} method, and 3) \emph{robust temporal median filter} towards pixel restoration based on low-rank subspace optimization theory. Fig.~\ref{fig:pipeline} illustrates the algorithmic pipeline of our proposed method.

As an initial study of an emerging topic, we focus on the case of static scenes. The problem setting, especially the hypothesis about the scenes and camera motions, will be in detail addressed in Section~\ref{sec:overview}. We target a fully automated method for solving this problem, unlike those works taking users in the loop~\cite{jian05,park10}. The extensions of the proposed method to other challenging settings (\eg, dynamically-moving objects in the scene) are of great important for practical consumer videos, which we keep as future exploration but discuss the possible solutions (\eg, motion layer analysis) in Section~\ref{sec:conclusion}.

%However, the framework can be extended to support other cases by coupling motion layer analysis.

%\begin{figure}[t!]
%\centering
%\subfigure{
%  \includegraphics[width=0.45\linewidth]{figures/flower_15.jpg}
%  }
%  \subfigure{
%  \includegraphics[width=0.45\linewidth]{figures/football_34.jpg}
%  }
%  \subfigure{
%  \includegraphics[width=0.45\linewidth]{figures/flower_15_processed.jpg}
%  }
%  \subfigure{
%  \includegraphics[width=0.45\linewidth]{figures/football_34_processed.jpg}
%  }
%\caption{\footnotesize Illustration of video de-fencing. The top row contain two frames selected from fence-corrupted video clips ``Flower" and ``Football" respectively. The bottom row presents the video restoration results using our proposed method. All the operations (including fence detection and removal) are fully automatic.}
%\label{fig:example}
%\end{figure}

\section{Related Works}

\subsection{Video Editing and Composition}

A vast literature has been devoted to video editing or composition in the past decade, such as semantic object cutout~\cite{baixue09}, motion magnification~\cite{liuce05}, video stabilization~\cite{yasu05}, video matching~\cite{sand04} and video completion~\cite{wexler07}. An interesting topic in this field is the removal of characteristic structures, such as rains or snowflakes. For example, Garg et al.~\cite{garg04} treated the visual manifestations of rain as a combination of both the dynamics of rain and the photometry of the environment. A correlation model that captures the dynamics of rain and a physical motion model that explains the photometry of rain are coupled for detecting and removing rain from videos. The work in~\cite{liuyanxi08} investigated semi-automatic fence detection and removal for images. However, the de-fencing quality therein is heavily dependent on the availability of repeating textures. The work in~\cite{park10} later extends the method to multi-view images, which is the most relevant work to the proposed method in this paper. However, the multi-view input therein are only loosely correlated and no strict temporal consistency is enforced. Heterogeneous views of the same scenes are used to find matching patches from the score of SSD (sum of squared difference) between local patches, rather than from temporal alignment and parallax cues. In this sense, Park et al.~\cite{park10} failed to provide an integrated framework to utilize visual parallax for video de-fencing.

Major difficulties in this field stem from the challenging problems like depth estimation and point correspondence. In image stitching and panorama generation from
multiple images or videos, correspondence can be obtained using invariant features~\cite{brown07}. However, as stated in~\cite{szeliki06}, parallax caused by depth significantly degrades the final quality yet has not been adequately solved. Various methods have been proposed to estimate depth~\cite{saxena08,zhangguofeng09}.

\subsection{Robust Data Recovery}

Prior work on video completion~\cite{shiratori06,wexler07} treats video as spatio-temporal cube. Various diffusion procedures are utilized to fill the missing parts in the videos. As will be shown later, our proposed method adopts a different idea. It basically hinges on non-parametric parallax-aware frame alignment, which also differs from geometric reconstruction based methods~\cite{zenggang07}. Pixel restoration on aligned frames boils down to robust data recovery in the existence of arbitrarily-corrupted outliers. Median filter is widely adopted for this task owing to its empirical success. Many variants have also been proposed, \eg, weighted median filter (WMF)~\cite{brownrigg}.

Median filter operates on scalars. For images or videos, the spatial smoothness or temporal coherence provides contextual regularization to mitigate the adverse effect of outliers. Since high-dimensional representation (vector, matrix, or tensor) is more natural for visual data, matrix-based robust recovery has recently attracted increasing attention. Recent endeavor on visual recovery borrows tools from the sparse learning~\cite{wright09b} and optimization theory. A representative work can be found in~\cite{wright09}, wherein several exemplar applications are presented like face recovery and surveillance background modeling.

\section{Overview}
\label{sec:overview}

\begin{figure}[thb]
\centering
\includegraphics[width=\linewidth]{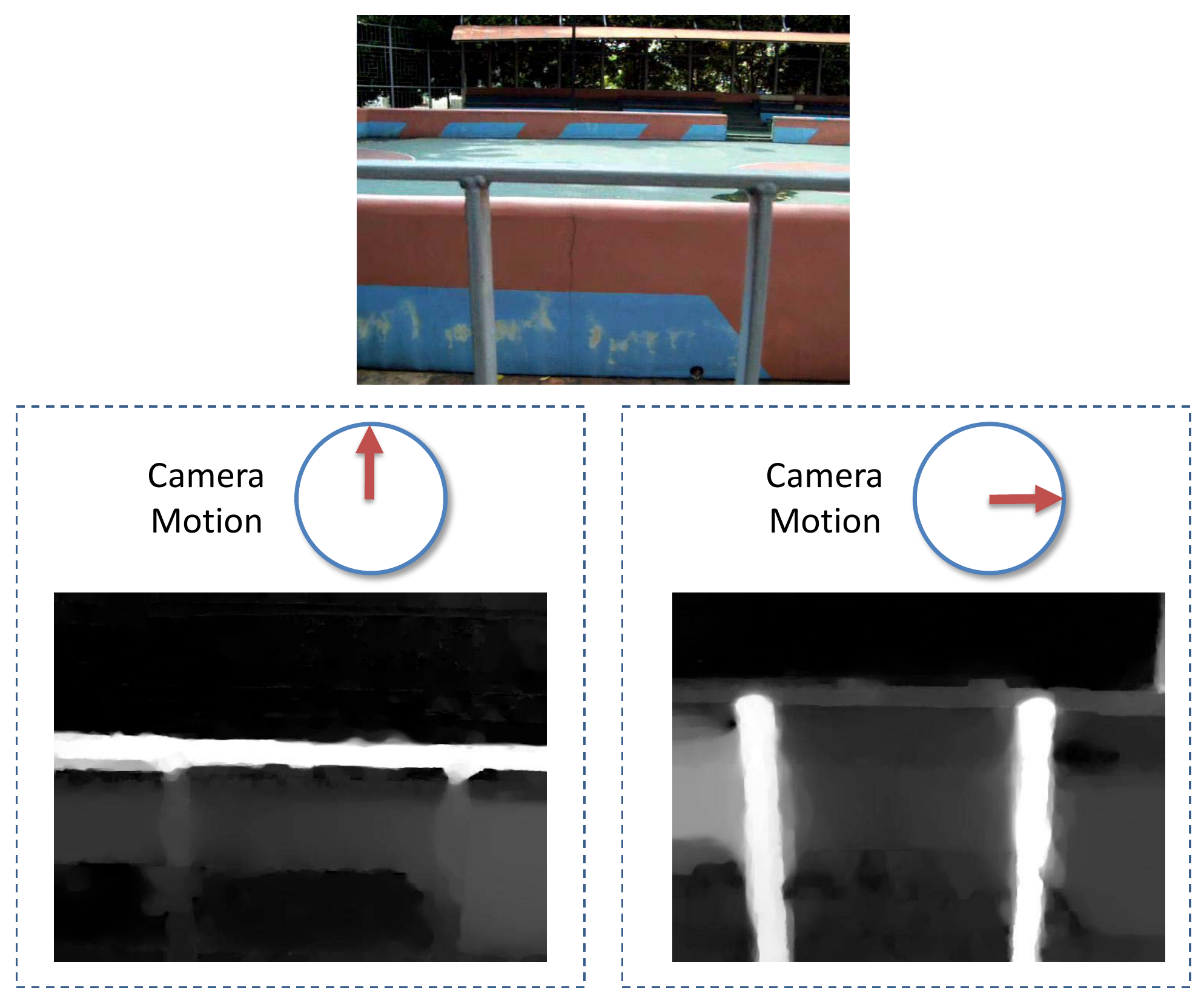}
\caption{\footnotesize The desired camera motion is related to the geometry of the fences. This example illustrates the effect of two kinds of camera motions (\ie, vertical and horizontal motions respectively) on the ``T"-shaped fence.}
\label{fig:camera}
\end{figure}

\begin{figure}[thb]
\centering
\includegraphics[width=\linewidth]{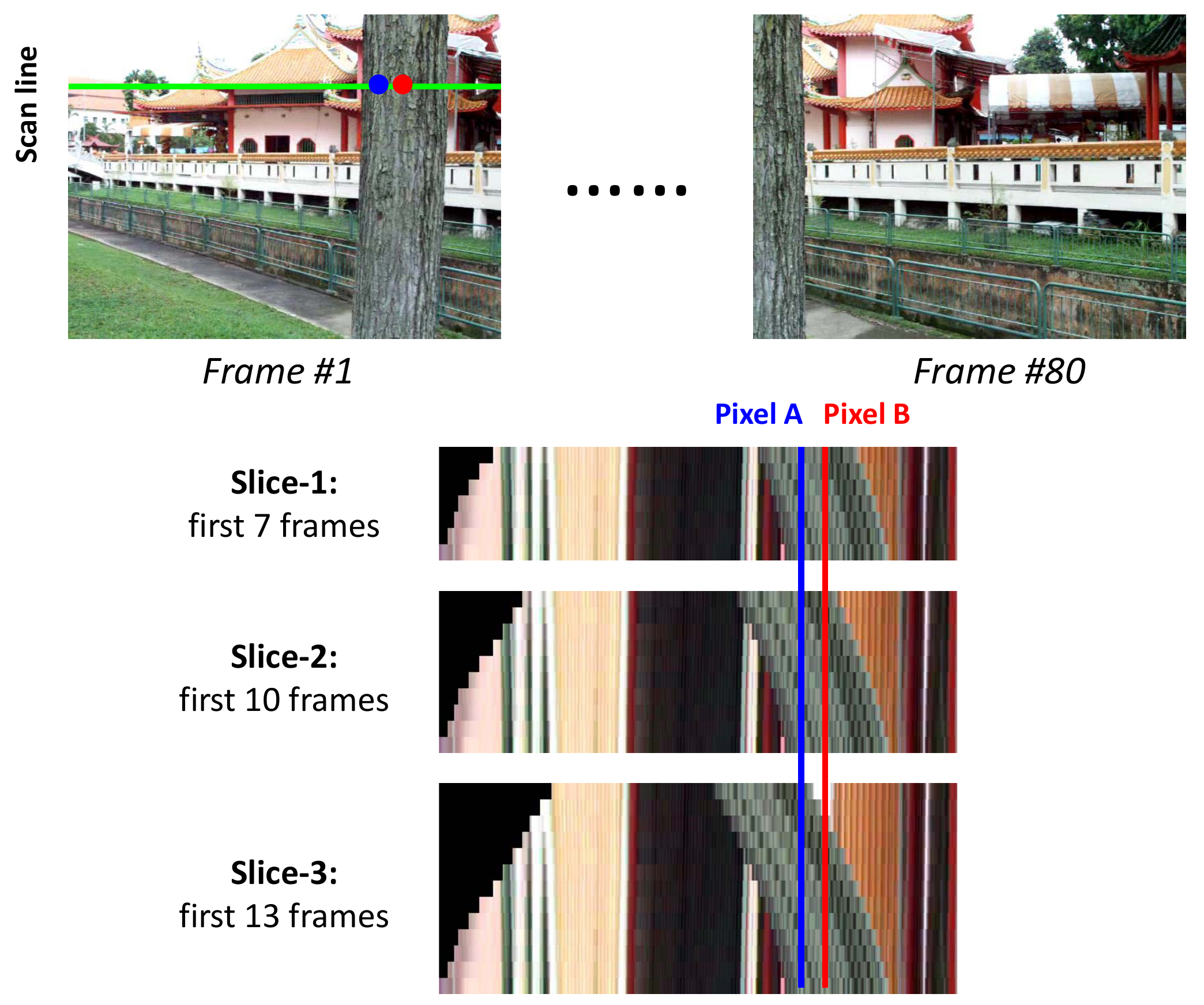}
\caption{\footnotesize Illustration of the restoration of occluded pixels. The top row shows a consecutive video frame sequence. Here the goal is to remove the occluding tree from the scene. The green line denotes an arbitrary scan line in the first frame along the horizontal direction. Putting the first $K$ frames (K = 7, 10, 13 respectively in this example) together with scene pixel aligned (see Section~\ref{sec:pof} for more technique details), it is possible to get image slices along the scan line. It is intuitively observed that pixel A on the scan line cannot be correctly restored since it is occluded for all $K$ values, and pixel B can be restored only for $K=13$. Best viewed in color.}
\label{fig:camera2}
\end{figure}

Our basic observation is that pixels occluded by the fences in a frame tend to become un-occluded along with the camera motion. In other words, suppose multiple consecutive frames are carefully aligned in pixel-by-pixel manner, ensuring in most cases the identically coordinated pixels along the temporal axis correspond to the same semantic objects except for the occluded pixels. The occlusion by fences can be consequently eliminated via pixel substitution from the relevant frames which are unobstructed from the corresponding viewpoint. Fig.~\ref{fig:model} illustrates the principal mechanism for solving the video de-fencing problem, highlighting the parallax resulting from disparate scene depths. As in prior exposition, we make reasonable assumptions for both achieving a solvable problem and covering a wide ranges of consumer videography. Specifically, the most important assumptions are listed as below:
\begin{itemize}
\item Fence-like occlusions are overwhelmingly closer to the camera compared with the target scene. Note that here the term ``fence" generally refers to anything excluded from the target scene, like an object right behind or in front of the real fence. It is also expected that the fence has a long, thin shape. As will be shown in the experiments, such shapes benefits more robust and exact video restoration.
\item The moving path of the camera is approximately parallel to the planes of fences. The goal is to make two consecutive frames approximately undergo affine transform, avoiding untractable perspective distortion. Moreover, the ideal camera moving direction is heavily dependent on the geometry of the fences. An example is shown in Fig.~\ref{fig:camera}. Due to the special ``T"-shaped fence, neither vertical nor horizontal camera motion is able to generate the expected visual parallax. Only an in-between camera motion can capture all branches of the fence. Finally, the magnitude of camera motion should be salient enough to guarantee the un-occlusion of every part of the scene in a number of consecutive frames. Fig.~\ref{fig:camera2} shows an example, where it is observed that some pixels are impossible to be restored under reasonable parameters. It is due to two reasons, either thick fence (\eg, the tree in Fig.~\ref{fig:camera2}) or weak camera motion, both of which should be taken into account during video capture.
\end{itemize}

The above rules imply that the proposed method is not applicable for general scenes. However, the rules cover a large spectrum of short-duration consumer videos (typically lasting only several seconds) and a user becomes qualified to capture the desired video clips after simple training.

We term this task as ``video de-fencing", which further boils down to two sub-tasks, \ie, \emph{probability-of-fence} (PoF) estimation and pixel selection. The former refers to the identification of those pixels undergoing fence occlusions, and the latter task tries to restore the visual information of high PoF pixels from temporally neighboring frames. Due to the ambiguity in pixel correspondence and complex scene structure, both sub-tasks are known to be difficult. The following sections address these sub-problems respectively.

%When the camera moves, parallax caused by the depth discrepancy between fence and scene can be observed. Assume the scene are remote enough to the camera and the transform between consecutive frames is approximately affine. After pixel-wise alignment, the image regions corresponding to fences are usually notably shifted in the image space, as seen in Figure~\ref{fig:model}.

\section{Probability-of-Fence (PoF) Estimation}
\label{sec:pof}

\begin{figure}[thb]
\centering
\includegraphics[width=0.47\linewidth]{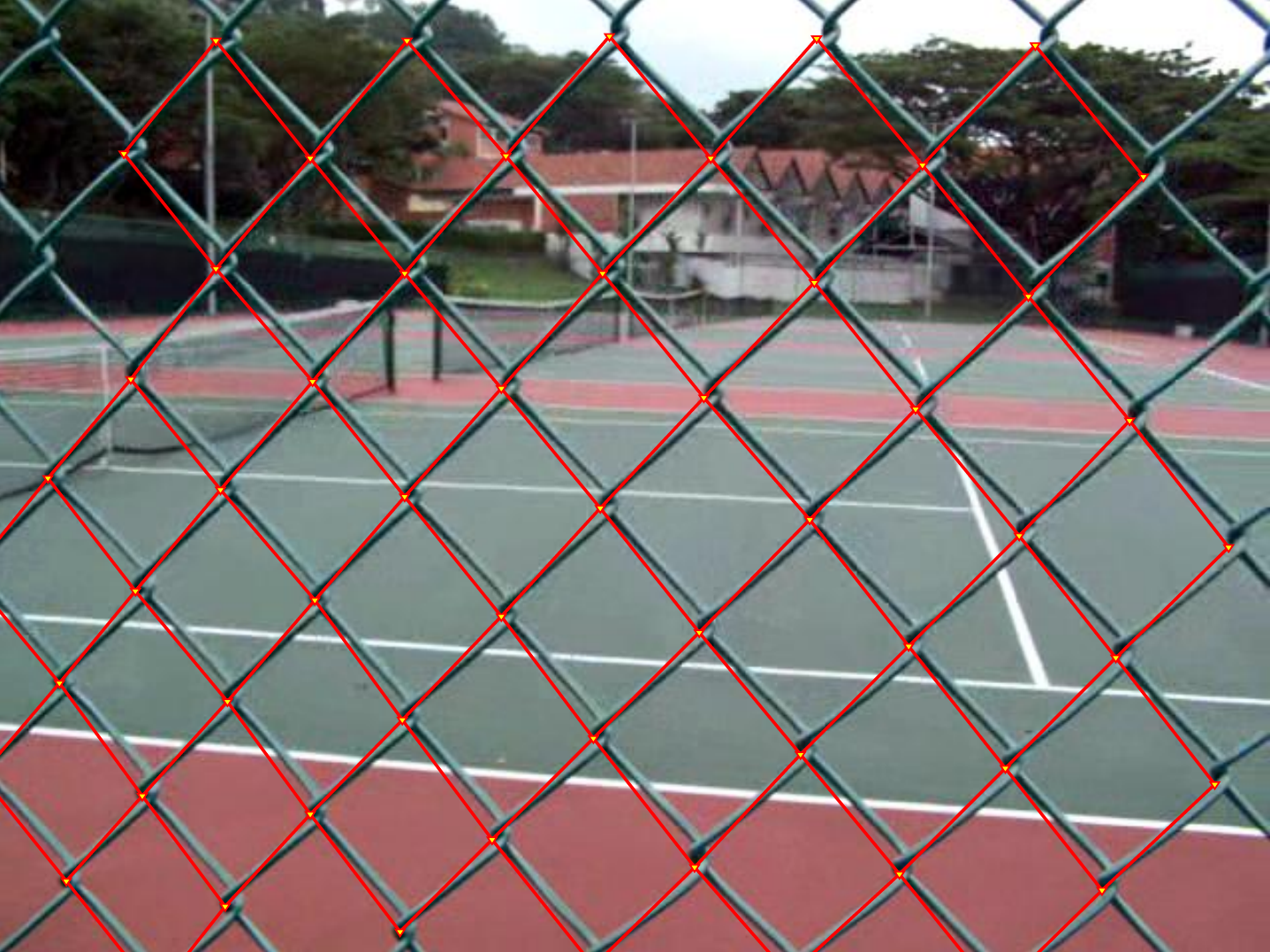}
\includegraphics[width=0.47\linewidth]{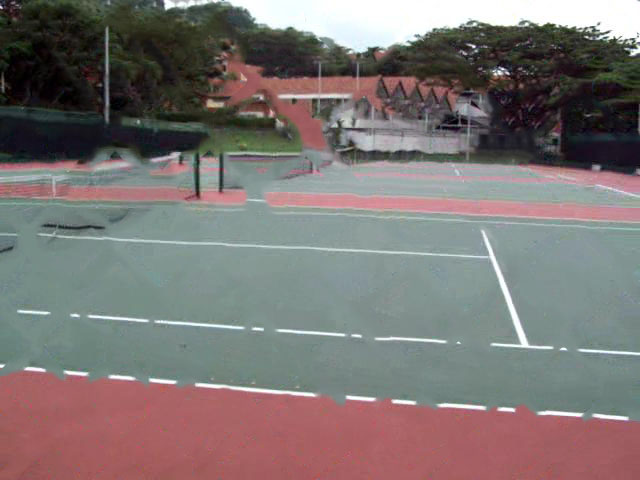}
\caption{\footnotesize The left figure draws the result using the state-of-the-art fence detector~\cite{liuyanxi08}. Several fence grids fail to be detected. It also tend to fail on non-repeating fences. The right figure shows the image completion result (with manually-labeled fence mask) using the build-in function in PhotoShop CS5.}
\label{fig:fencedetector}
\end{figure}

The goal of this stage is to estimate the confidence value of each pixel coming from fences. Before proceeding, we first introduce two existing solutions towards this goal:
\begin{itemize}
\item In prior study ``image de-fencing"~\cite{liuyanxi08}, fences are assumed to have visual regularity such as structural symmetry. However, our empirical investigation reveals its inapplicability. The diversity of fence appearance (see Fig.~\ref{fig:data}) can be hardly encompassed using simple visual rules like symmetry or local low-rank texture~\cite{zhang10}. Even state-of-the-art fence detector need be improved to be more efficient and practical for real-world video sequences. Fig.~\ref{fig:fencedetector} provides a failure example on the video clip ``Tennis" (the left sub-figure). Fig.~\ref{fig:data} further presents some video clips (\eg, ``WinterPalace" and ``Temple") that do not contain any symmetry, which complicates the fence detection.
\item It is another natural solution to manually specify an image mask and resort to image completion algorithms~\cite{iddo03,sunjian05}. Fig.~\ref{fig:fencedetector} presents the results obtained by the ``content-aware fill" function implemented in commercial software PhotoShop CS5. This new feature helps the users retouch any image in removing unwanted areas. It is able to do it by filling the space by utilizing pixels which surround it. However, occlusion is generally un-recoverable from a single frame.
\end{itemize}

Unlike previous work, we perform \emph{probability-of-fence} (PoF) estimation based on visual parallax. Two types of cues are utilized to infer the PoF values, \ie, visual flow analysis and image appearance differencing depicted in Sections~\ref{subsec:flow} and~\ref{subsec:app} respectively. Incorporating the motion cue from visual flow analysis reflects the assumption that scene background and fence-like objects can be effectively distinguished from parallax. However, in some cases, due to the ambiguity in the motion estimation, motion cue can be noisy (see Fig.~\ref{fig:flow}(a) for an example), which motivates us to utilize appearance cue. The basic idea is to first estimate the mode of the motions within two consecutive frames, and then perform frame alignment accordingly. After differencing aligned frames, those pixels with large appearance variances will be assigned high probabilities of being from fences, since their motion vectors probably deviate a lot from the motion mode. However, as shown in Fig.~\ref{fig:flow}(c)(d), appearance cue tends to generate undesired striped PoF patterns for uniform-colored fences (since the response is only strong at the edge of the fences), which indicates that these two kinds of cues are indeed complementary to each other in many cases.

Due to the video noises and the inconsistency between the presumed motion models and real-world scene geometry, making binary decision per pixel (\ie, coming from either fence or target scene) is error-prone. Instead, each pixel is assigned soft confidence value in $[0,1]$, indicating the probability to be from the fence. The final PoF confidence values are obtained via the linear combination of these complementary information channels.

\subsection{Motion Cue}
\label{subsec:flow}

\begin{figure}[thb]
\centering
%\subfigure[Motion based PoF]{
%\includegraphics[width=0.46\linewidth]{./figures/tennis_20_flow.pdf}
%}
%\subfigure[Appearance based PoF]{
%\includegraphics[width=0.46\linewidth]{./figures/tennis_20_app.pdf}
%}
%\subfigure[Motion based PoF]{
%\includegraphics[width=0.46\linewidth]{./figures/football_24_flow.pdf}
%}
%\subfigure[Appearance based PoF]{
%\includegraphics[width=0.46\linewidth]{./figures/football_24_app.pdf}
%}
\includegraphics[width=0.95\linewidth]{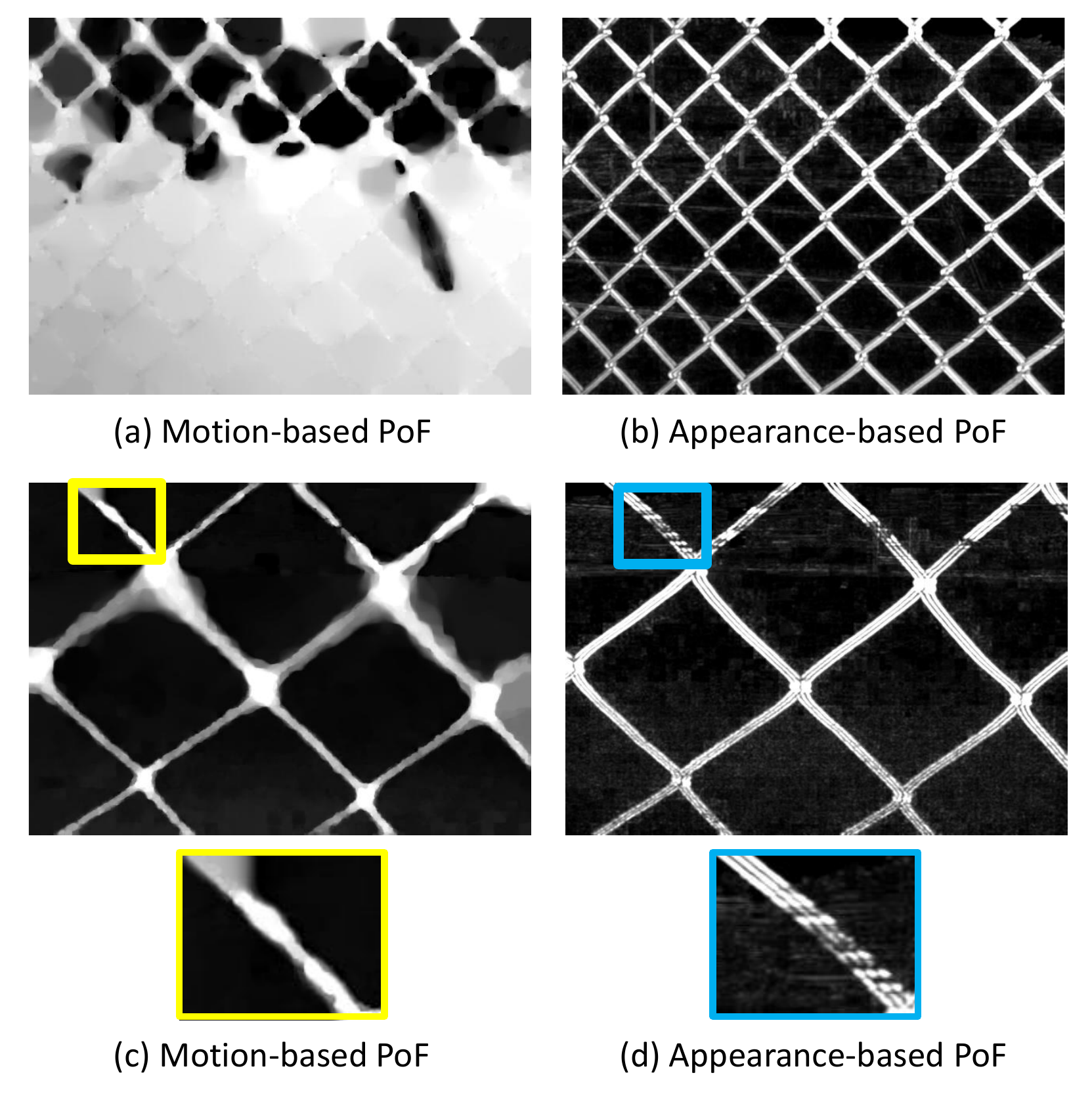}
\caption{\footnotesize PoF estimations for the 20-th frame in video clip ``Tennis" (see Fig.~\ref{fig:fencedetector} for an exemplar frame from this video clip) and the 24-th frame in video clip ``Football", shown in the top row and bottom row respecively.  Note that the motion cue and appearance cue tend to be complementary to each other. See text for more explanation.}
\label{fig:flow}
\end{figure}

%In (a) we adopt the visualization scheme in~\cite{liuce09}, where different hue indicates the direction of the flow vector and the saturation corresponds to the motion magnitude.

Since the video clips are assumed to be continuously captured, each frame can be reasonably mapped to the coordinate system of consecutive frames
according to tiny motion flows (usually fewer than 3-pixel displacement). However, the motion magnitude of fence pixels are notably larger due to the parallax phenomena. Consequently, it provides the possibility to judge the fence by distinguishing the saliently-shifted pixels.

The optical flow method~\cite{bruhn05,liuce09} is employed towards the aforementioned aim. Suppose from frame $F^{t}$ to frame $F^{t+1}$ it undergos a motion field valued as $w(p) = \langle u, v \rangle$ at the pixel with index $p$, where $u$ and $v$ denote the horizontal or vertical flow respectively. Under the Lambertian surface and brightness constancy assumptions, and the piecewise smoothness prior, we adopt the following objective function as in~\cite{liuce09} to guide the motion field optimization, \ie,
\begin{eqnarray}
\mathcal{J}(u,v)= \int_p \psi \left( | F(p+w) - F(p) | \right) + \alpha \phi(|\nabla u|^2+|\nabla v|^2),\nonumber
\end{eqnarray}
where $\psi(\cdot)$ and $\phi(\cdot)$ are robust norms taking the forms $\psi(x)=\sqrt{x^2+\epsilon^2}$ and
$\phi(x)=\sqrt{x^2+\epsilon^2}$ ($\epsilon$ is a small constant for numerical stability). $| \nabla u |^2 = u_x^2 + u_y^2$
penalizes large motion gradients ($u_x = \partial u / \partial x$, $u_y = \partial u / \partial y$). The high non-convexity of $\mathcal{J}(u,v)$ makes the optimization easily trapped in local optima. To address this issue, Gaussian image pyramid is constructed. The optimization initially starts from low-resolution image levels and then propagates to finer levels by bilinear interpolation. Following the work in~\cite{liuce09}, we calculate the first-order Taylor expansion of $\mathcal{J}(u,v)$ and adopt the \emph{iterative reweighted least squares} (IRLS) scheme for updating, which demonstrates high efficiecy (roughly 2 seconds for a $480 \times 240$-pixel image). Fig.~\ref{fig:flow} shows an example.

After obtaining the motion fields, it is necessary to compute its principal direction to abandon redundant information, which can be trivially computed from the covariance matrix, \ie,
\begin{eqnarray}
C = \mathbb{E}_p \left[ (w_p - \bar w)^T (w_p - \bar w) \right],
\end{eqnarray}
where $\bar w$ is the averaged motion vector. The principal direction (equivalent to the camera motion direction) is known to be the eigen-vector of matrix $C$ associated with the maximal eigen-value. Denote it to be $c$. All motion vectors are then projected onto $c$ to reduce redundant components\footnote{Note that both $c$ and $-c$ equivalently play the role of principal eigen-vector. We adopt the one which ensures $\mathbb{E} (c^T \bar w ) > 0$, forcing
the fence pixels to have larger positive projection values.}, obtaining 1-D scalar $m(p) = c^T \cdot w(p)$ for pixel $p$. The final confidence of fence-ness is calculated by choosing two thresholds $\theta_l$ and $\theta_h$ and performing a linear mapping as below:
\begin{eqnarray}
f_m(p) \leftarrow \frac{ \tilde m(p) - \theta_l } { \theta_h - \theta_l} \in [0,1],
\end{eqnarray}
where $\theta_h > \theta_l$ and $\tilde m(p) = \max(\min(m(p),\theta_h),\theta_l)$ for robustness consideration. In implementation, both thresholds are generated in data-driven manner, \ie, $\theta_h$, $\theta_l$ are chosen to be $90\%$, $10\%$ largest values among all the projected values. For frame $F^t$, it has bidirectional mapping (\ie, to frame $F^{t-1}$ or frame $F^{t+1}$). We compute the confidence value in each direction and takes the averaged value as the final result. See Fig.~\ref{fig:flow} for an example.

\subsection{Appearance Cue}
\label{subsec:app}

The optical flow method is often inaccurate due to the aperture problem and corrupted pixels during capture. We empirically find that appearance cue complements above-mentioned motion cue in many cases (see Fig.~\ref{fig:flow} for an example). Specifically, we assume that frame $F^t$ is mapped to frame $F^{t+1}$ by parametric affine transform, \ie, the original coordinate $\langle x, y \rangle$ is projected to the new one $\langle x', y' \rangle$ via $x' = a_1 x + a_2 y + a_3$, $y' = a_4 x + a_5 y + a_6$, where $a_1,\ldots,a_6$ are the coefficients to be optimized. At least three correspondences are required to estimate these six parameters.

For this aim, we utilize the local keypoint based image alignment algorithm~\cite{szeliki06}. On each video frame, SIFT features are extracted and matched between consecutive frames. Since the coordinates of SIFT features in frames $F^t$ and $F^{t+1}$ are known, the six parameters can be reliably estimated from the SIFT correspondences by least-squares. Note that the fences and scene always undergo different affine transforms due to the visual parallax. To enhance the robustness, we further modulate each pixel by their motion-induced PoF value $f_m(p)$. The SIFT feature with high $f_m(p)$ values will be assigned low weights (in practice we adopt 1-$f_m(p)$), resulting a weighted least-squares estimator.

%The procedure iterates between $F^t \rightarrow F^{t+1}$ and $F^t \rightarrow F^{t-1}$. Eventually an $h \times w \times 3$ spatial-temporal visual cube is obtained, where $h$, $w$ correspond to the image height and width respectively.

With the estimated affine transform, the temporally-adjacent frames can be accordingly aligned to a specific frame and thereby it enables the analysis of any pixel $p$ on this frame using the geometrically-aligned spatio-temporal cube. We can get the property of a pixel $p$ by analyzing a spatial-temporal patch around it (in practice we adopt the size $5 \times 5 \times 3$). Various operators have been proposed to estimate the regularity of such spatial-temporal structures, \eg, the eigen-spectrum based motion estimator~\cite{shechtman07}. Intuitively, any aligned pixel tends to be from target scene if it has small intensity variation along the temporal dimension (see the frame slices in Fig.~\ref{fig:camera2}). For numerical stability, we regularize it using the variation along the $(x,y)$ dimensions. For each pixel, its variations along the $(x,y)$ image plane and temporal dimension are estimated and denoted as $\sigma_{xy}(p)$, $\sigma_t(p)$ respectively. The appearance-induced fence-ness is defined as below:
\begin{eqnarray}
f_a(p) \leftarrow \tanh \left( \frac{ \sigma_t (p)} { 2 \cdot \gamma + \sigma_{xy}(p) } \right) \in [0,1],
\label{eqn:fa}
\end{eqnarray}
where $\gamma = \mathbb{E}_p (\sigma_{xy} (p))$ is introduced to penalize small $\sigma_{xy}(p)$ in uniform image regions, and $\tanh(\cdot)$ is a function that maps the input scalar to the range $[0,1]$.

\subsection{Bundle Adjustment}

The final PoF value is computed by linearly combining $f_m(p)$ and $f_a(p)$, \ie, $f(p) = \lambda f_m(p) + (1-\lambda) f_a(p)$. After the computation over all frames, bundle adjustment can be adopted for further noise suppression. Let $\mathcal{N}(p)$ be the index set of neighbors (either in spatial or temporal scale) for pixel $p$, its PoF value is updated according to
\begin{eqnarray}
f(p) \leftarrow \kappa \sum_{q \in \mathcal{N}(p)} w_{pq} f(q) / \sum_{q \in \mathcal{N}(p)} w_{pq} + (1-\kappa) f(p),
\end{eqnarray}
where $\kappa$ is a free parameter to control the resistent strength to the incoming information. Given a well-defined neighborhood system, the above procedure is known to have convergence guarantee~\cite{jingdong09}.

\section{Pixel Restoration}

The next crucial step is to perform restoration on the occluded pixels (\ie, those associated with high PoF values). Unlike texture-based inpainting in image de-fencing, the video de-fencing capitalizes on temporal consistency of the pixels in the geometrically-aligned frame sequences. If the true color of a pixel has ever been exposed to the camera on partial frames, it is theoretically recoverable. There are two challenges that remain towards the ultimate goal, \ie, sub-pixel frame alignment and afterwards robust temporal filtering, which are detailed in Sections~\ref{subsec:alignment} and~\ref{subsec:rtmf} respectively.

A naive solution is first performing frame alignment using the global affine transform learned by the method in Section~\ref{subsec:app}, followed by temporal median filtering to restore occluded pixels. However, this idea practically suffers from several factors on the video clips captured by hand-held cameras. On one hand, the method in Section~\ref{subsec:app} cannot achieve sub-pixel image alignment under large occlusions and complicated scene depth structure. On other hand, \emph{naive temporal median filter} (N-TMF) only works when the ``correct" pixels dominate in quantity (see Fig.~\ref{fig:camera2}, where pixel B on slice-3 is theoretically recoverable yet the recovery will fail via N-TMF).

To remedy these problems, Section~\ref{subsec:alignment} elaborates on an occlusion-resistent image alignment algorithm based on truncated optical flow computation, wherein both sub-pixel accuracy and robustness to small scene motions are feasible. Moreover, we also propose \emph{robust temporal median filter} (R-TMF) in Section~\ref{subsec:rtmf}, which is possible to return the correct value even in the case that fence pixels dominate the pixel collection in quantity.

%Recall that the large motions (\eg, fences) tend to distort the flow field, deteriorating subsequent pixel restoration. Our method mitigates this adverse effect by pixel weighting and large-flow truncation.

%Moreover, we also propose \emph{robust temporal median filter} (R-TMF), which is possible to return the correct value even in the case that fence pixels dominate the pixel collection in quantity. Specifically, our basic idea is to model the frame appearances as the linear addition of a low-rank component (approximately aligned background scenes) and sparse arbitrarily-large outliers (\ie, fence pixels), principally following the work in~\cite{wright09}. These two components can be optimally separated by efficient convex optimization. More details are postponed in Section~\ref{subsec:rtmf}.

\subsection{Parallax-aware sub-pixel frame alignment}
\label{subsec:alignment}

\begin{figure}[thb]
\centering
\subfigure[Slice at $y=100$]{
\includegraphics[width=0.96\linewidth]{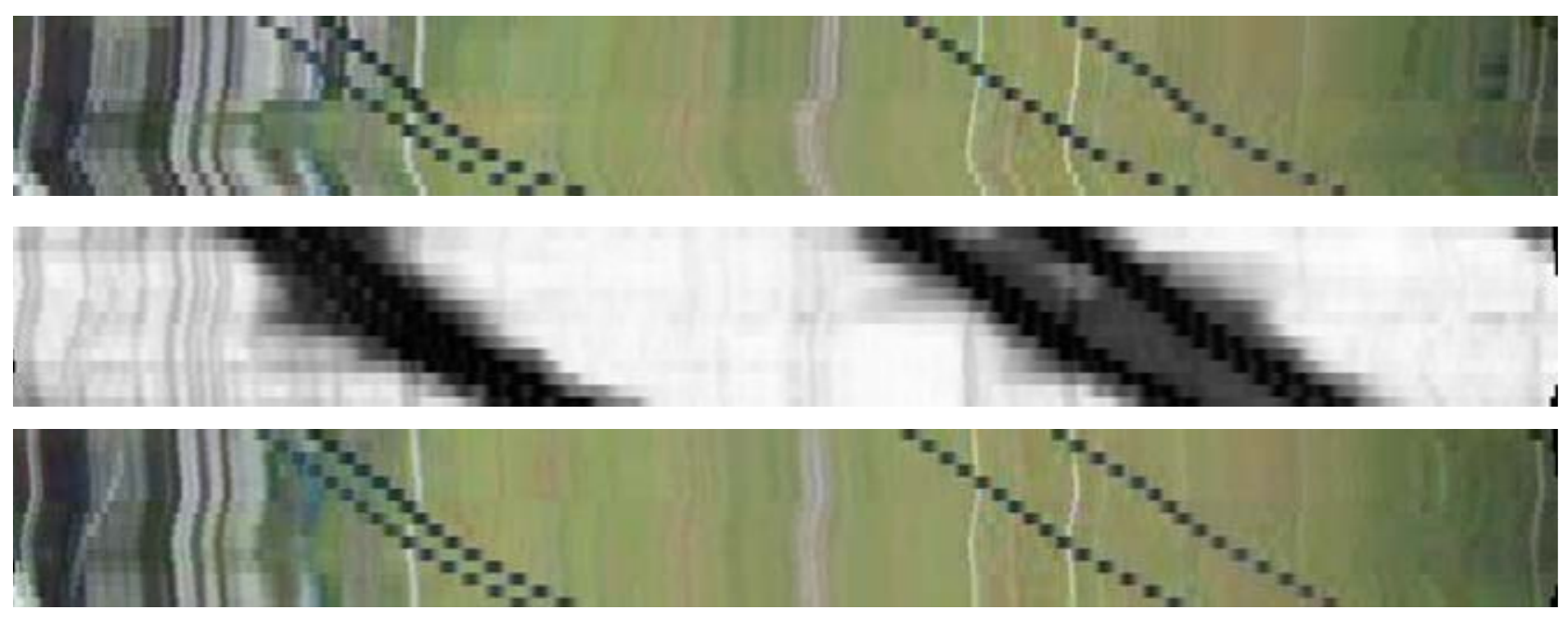}
}
\subfigure[Slice at $x=100$]{
\includegraphics[width=0.72\linewidth]{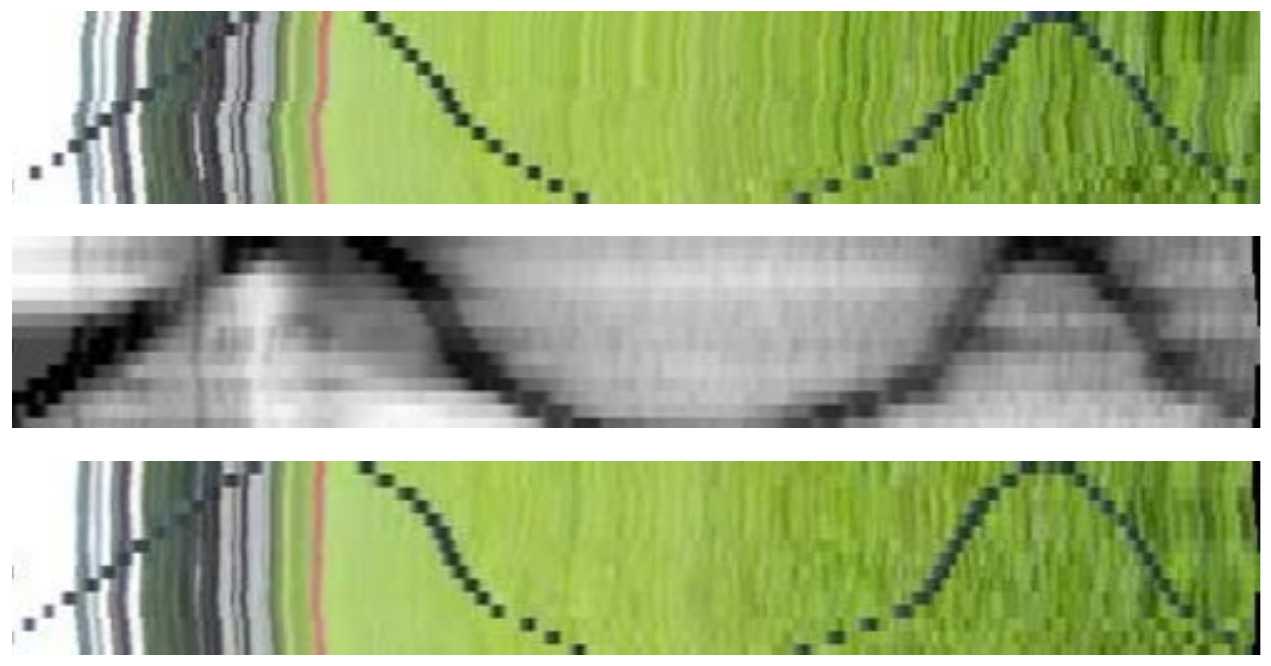}
}
\caption{\footnotesize Exemplar results of the proposed alignment algorithm. The spatial-temporal cube is centered at the 25-th frame in clip ``Football" (the original frame is seen in Fig.~\ref{fig:trunc}). (a) and (b) plot the slices along specific horizontal ($y=100$) or vertical ($x=100$) position respectively. The meanings of three slices from top to bottom: those before sub-pixel alignment, aligned fence-ness values, and those after sub-pixel accurate alignment.}
\label{fig:alignment}
\end{figure}

%sub-pixel image alignment. parallax-resistent image alignment.

To ensure all pixels observed in partial of the frames, the cardinality of the frame set used for pixel restoration is required to be large enough. For frame $F^t$, we choose the previous $M$ frames $\{ F^{t-M}, F^{t-M+1}, \ldots, F^{t-1}\}$ together with the next $M$ frames $\{ F^{t+1}, F^{t+2}, \ldots, F^{t+M}\}$ as the working set. As pre-processing, these frames are roughly aligned according to the transform matrix learnt using the method in Section~\ref{subsec:app}. Note that there is a dilemma to select proper value of parameter $M$. Large $M$ tends to convey more useful information yet complicates frame alignment. In practice we find that setting $M=7$ is proper for most video sequences.

Our initial study reveals the incapability of assuming holistic or block-varying geometry transform (\eg, affine or projective), which tends to produce over-blurry results and discard thin scene objects like flagpoles due to the accumulated misalignment within the $2 M + 1$ frames. In practice sub-pixel accuracy is required to guarantee the performance. However, the algorithm will fail in the case that fences are also aligned, which nullifies the temporal cue used for pixel restoration. As stated above, fence or scene pixels are distinguished according to the motion magnitudes, \ie, the parallax. An ideal image alignment is expected to take parallax into account.

To address above issues, we propose a weighted, truncated optimal flow method. The objective function to be minimized can be expressed as:
\begin{eqnarray}
\label{eqn:motion}
\mathcal{J}^e (u,v) \triangleq \int_p \Big[ (1-f(p)) \cdot \psi \left( | F(p+w) - F(p) | \right) \nonumber \\
\quad + \alpha \cdot f(p) \cdot \phi(|\nabla u|^2+|\nabla v|^2) \Big],
\end{eqnarray}
under the following constraints:
\begin{eqnarray}
-\theta_u \le u \le \theta_u,  \\
 -\theta_v \le v \le \theta_v.
\end{eqnarray}

PoF values $\{f(p)\}$ are utilized to suppress fence-like pixels. Parameters $\theta_u$, $\theta_v$ are used to truncate salient motions probably from fences. The optimization described in (\ref{eqn:motion}) iterates between $F^t$ and each of its $2 M$ temporally-adjacent frames. For compensation of accumulated misalignment between $F^t$ and $F^{t+k}$, we empirically find $\theta_u = \theta_v = 1.5 k + 1$ works well on most videos. See Fig.~\ref{fig:alignment} for an example.

%However, the aforementioned pre-processing rough alignment mentioned in Section takes place between neighboring frames. This inconsistency indicates the accumulated misalignment between $F^t$ and $F^{t+k}$ is probably large when $k$ is increased.

\begin{figure}[t!]
\centering
\includegraphics[width=0.9\linewidth]{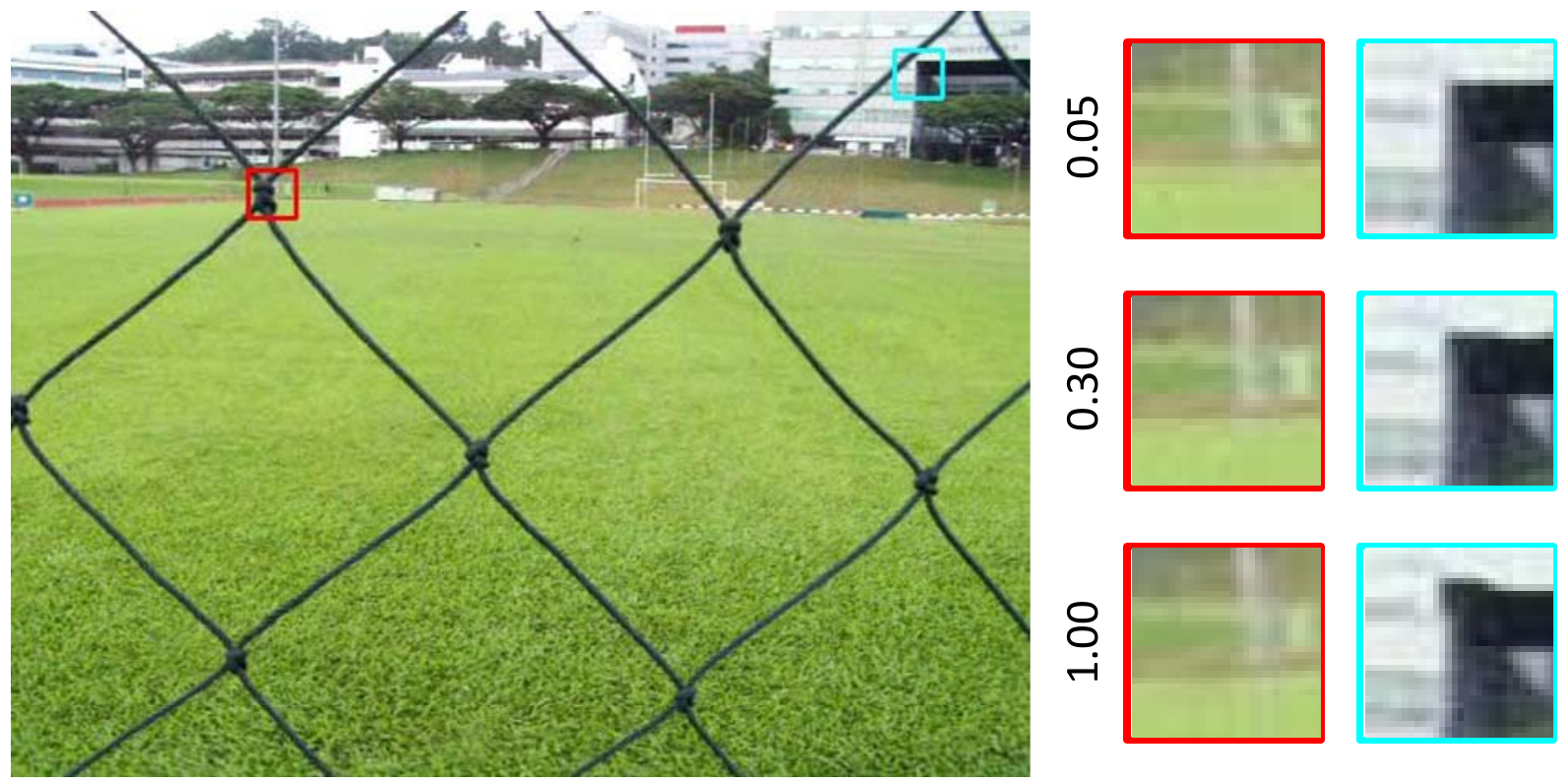}
\caption{\footnotesize The close-up views of restored frame under different $\eta$. It is observed that high $\eta$ tends to distort local image structures.}
\label{fig:trunc}
\end{figure}

%The original frame $F^{t+k}$ is warped to align $F^t$ according to the optimized motion vectors.

Recall that $\theta_u$, $\theta_v$ tend to be loose bounds. To avoid large motions distorting the overall motion field, another form of motion truncation is executed after computing the motion vectors. Particularly, pixels lie on different scan lines along the principal motion direction. The motion means and standard variations along each scan line are calculated, denoted as $\mu_m$ and $\sigma_m$ respectively. Afterwards all motions on this line are truncated to be within $[\mu_m - \eta \cdot \sigma_m, \mu_m + \eta \cdot \sigma_m]$. We empirically find that smaller $\eta$ (\eg, 0.1) produces reasonable results. See Fig.~\ref{fig:trunc} for an example.

\subsection{Robust Temporal Median Filter (R-TMF)}
\label{subsec:rtmf}

In the aligned $h \times w \times (2 M + 1)$ spatial-temporal cube, at most $2 M + 1$ pixels reside on each line orthogonal to the image plane. A direct solution to pixel restoration is applying naive temporal median filter (N-TMF) onto such pixel ensemble to resist the detrimental fence pixels. Unfortunately, for reliable restoration via N-TMF, it is crucial to ensure the ``correct" pixels dominates, which makes N-TMF unstable and fail in several cases, including small parallax, thick fences along the principal motion direction and under-estimated parameter $M$.

To address above issues, we propose the so-called \emph{robust temporal median filter} (R-TMF), which is possible to survive even when corrupted pixels dominate. The key idea is to weight the pixels with estimated PoF confidences, such that fence pixels are suppressed to gain more robustness. Moreover, the global luminance and chromatic statistics may be inconsistent between consecutive frames partially owing to automatic camera white balancing and environmental lighting change. Under this condition N-TMF is known to be sensitive. In contrast R-TMF implicitly models such inter-frame alteration based on low-rankness matrix prior.

Formally, given pixel collection $\{x_t\}_{t \in \mathcal{I}}$, robust estimator theory has disclosed the equivalence between the output of N-TMF and the minimizer of $\arg \min_\mu \sum_t \|x_t-\mu\|_1$. Our proposed R-TMF actually extends this observation into the vector case, wherein any $x_t$ is a vector rather than scalar as in N-TMF. In current context, $x_t$ denotes any equivalent vector representation of original image matrix $F^t$ (or tensor for multi-channel images). We use the notation $X = [x_1,\ldots, x_{2 M + 1}]$ to represent the 2-D data matrix, piling all $x_t$ as its column vectors. Mathematically, the goal of R-TMF is to decompose $X$ into two additive components, motivated by the \emph{robust principal component analysis} (R-PCA) framework in~\cite{wright09}:
\begin{eqnarray}
\label{eqn:rpca}
\min_{A,E} \parallel A \parallel_\ast + \lambda \parallel E \parallel_1 \quad s.t. ~~X = A + E,
\end{eqnarray}
where $\parallel \cdot \parallel_\ast$ denotes the matrix nuclear norm, returning the sum of its singular values. It is known as a widely-used convex surrogate for non-smooth matrix rank. $\parallel \cdot \parallel_1$ is matrix $\ell_1$-norm, returning the sum of the absolute of all matrix elements. Analogously, R-TMF generalizes the scalar mean in N-TMF to be matrix nuclear norm (both encourage simplicity), and scalar absolute to be matrix $\ell_1$-norm (both are robust to extremely-large outliers).

\begin{figure}[t!]
\centering
\includegraphics[width=0.85\linewidth]{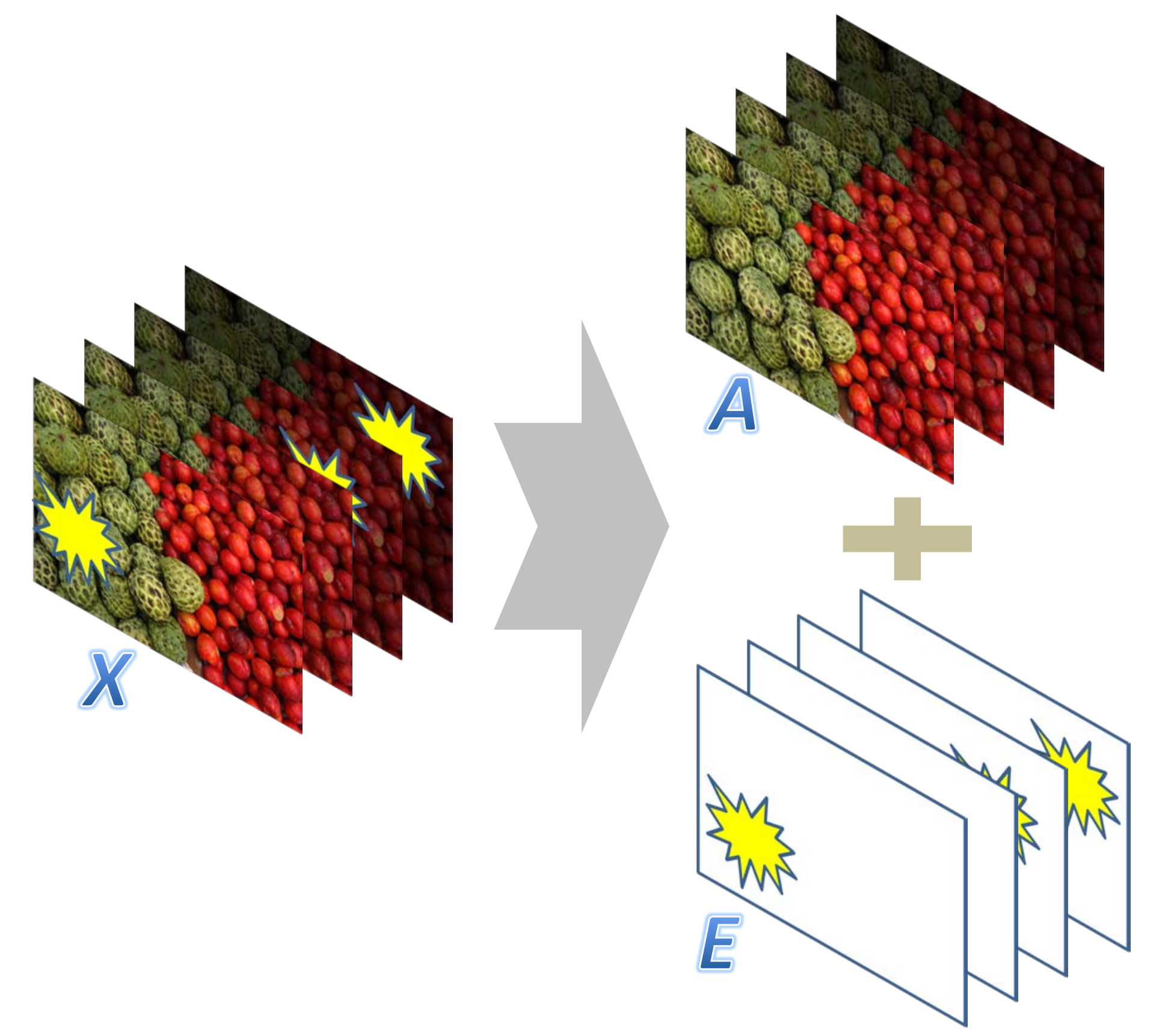}
\caption{\footnotesize Illustration of the basic idea of R-TMF.}
\label{fig:rtmf}
\end{figure}

\begin{figure*}[thb]
\centering
\subfigure[Original Frame]{
\includegraphics[width=0.28\linewidth]{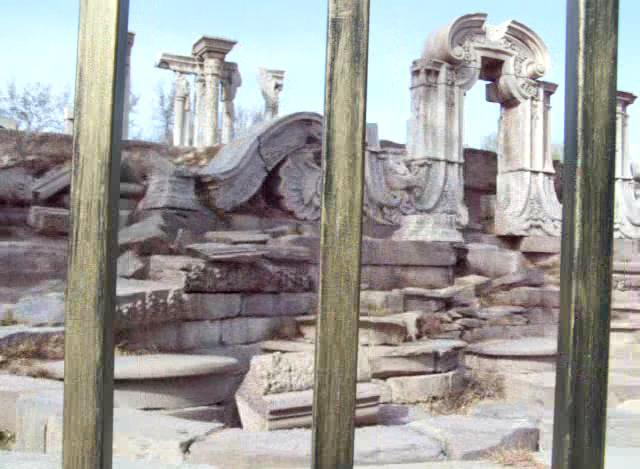}
}
\subfigure[Naive Median Filter]{
\includegraphics[width=0.28\linewidth]{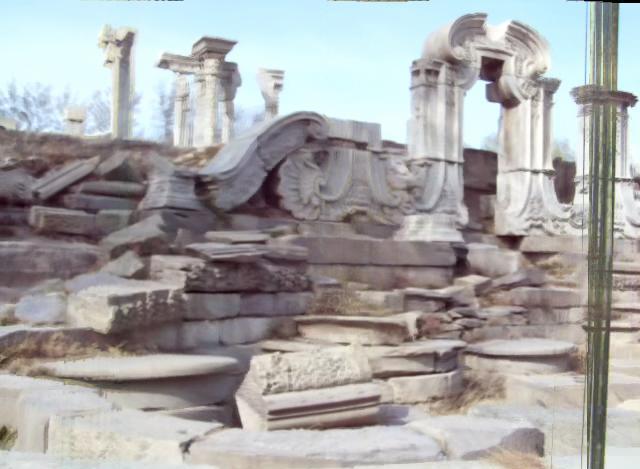}
}
\subfigure[Average Filter]{
\includegraphics[width=0.28\linewidth]{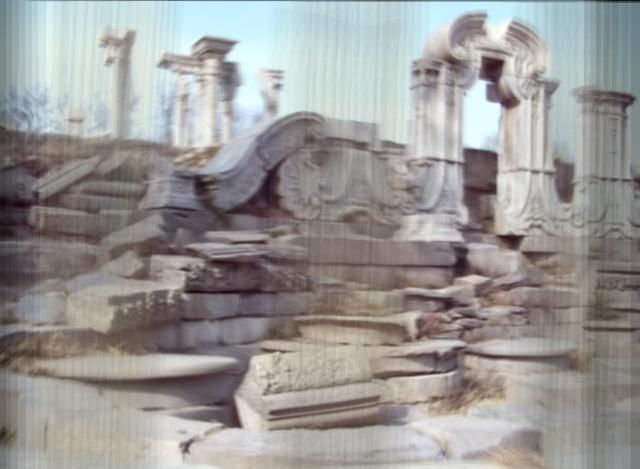}
}
\subfigure[Mask-based Image Completion]{
\includegraphics[width=0.28\linewidth]{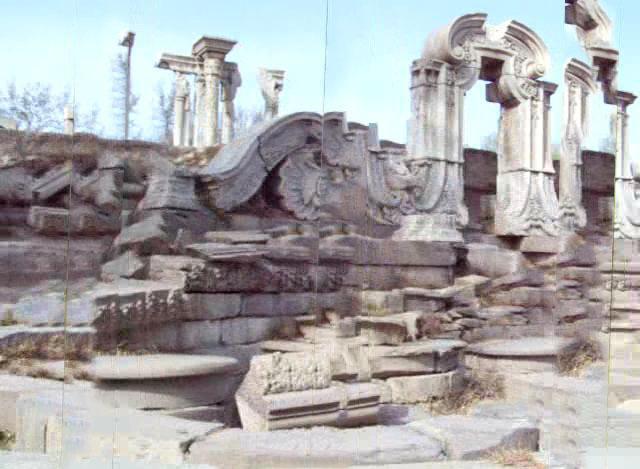}
}
\subfigure[Our Proposed Method]{
\includegraphics[width=0.28\linewidth]{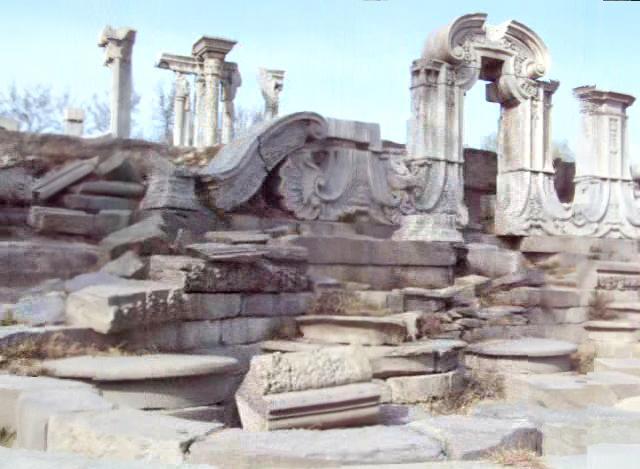}
}
\caption{\footnotesize Comparison of different estimators on the ``WinterPalace" video sequence.}
\label{fig:tmf2}
\end{figure*}

To better clarify the intuition underlying R-TMF, it is possible to factorize the low-rank component in (\ref{eqn:rpca}) as $A = P Q^T$, where $P = [p_1,\ldots,p_r]$, $Q = [q_1,\ldots,q_r]$ (assume $rank(A)=r$ without loss of generality). In the context of video de-fencing, $P$ is comprised of image bases and $Q$ conveys the information like camera parameters and environmental conditions. Fig.~\ref{fig:rtmf} illustrates it on toy data, where aligned images $X = [I^1, \ldots, I^L]$ are corrupted by moving shining blob. By optimizing (\ref{eqn:rpca}) the uncorrupted images and blobs can be separated. Moreover, we further assume the images undergoing a continuous luminance attenuation parameterized by $I^k = c^{k-1} I^1$ ($0 < c < 1$). In the ideal case, it is possible to find a factorization such that $P = I^1$, $Q = (c^0, c^1,\ldots,c^{L-1})^T$, and $rank(A) = rank(P Q^T) = 1$.

The formula in~(\ref{eqn:rpca}) is degraded to median filter when $\lambda \rightarrow \infty$. As a 2D extension of N-TMF, it still tends to fail under heavy outliers, which are common in the video de-fencing context (\ie, fence pixels dominate along many spatial-temporal directions in the aligned $2 M + 1$ frames). Directly solving~(\ref{eqn:rpca}) is difficult to obtain satisfactory results. A straightforward solution is to explicitly specify the weights for each element in matrix $X$, so that the adverse effect of the outliers will be mitigated. We choose the PoF information for this aim, and enhance~(\ref{eqn:rpca}) to obtain its weighted version:
%Recall that the PoF information is available, we can further enhance (\ref{eqn:rpca}) to obtain its weighted version:
\begin{eqnarray}
\label{eqn:rpca2}
\min_{A,E} && \parallel A \parallel_\ast + \lambda \parallel W \otimes E \parallel_1 \\
s.t. && W \otimes X = W \otimes A + W \otimes E,
\end{eqnarray}
where $\otimes$ is element-wise matrix product and $W$ represents the weight matrix with its $(k,p)$-th entry equal to PoF-induced value $1-f(p)$ in frame $k$. The problem in~(\ref{eqn:rpca2}) is convex, whose global optima can be efficiently pursued by convex solvers. We propose to utilize an efficient optimization algorithm which capitalizes on \emph{augmented Lagrange multipliers} (ALM)~\cite{bertsekas}.

For completeness, we first briefly introduce the basics of ALM and then sketch the algorithmic pipeline. In~\cite{bertsekas}, the general method of augmented Lagrange multipliers is introduced for solving constrained optimization problems of the kind:
\begin{eqnarray}
\min_X f(X), \quad s.t. ~~h(X) = 0,
\label{eqn:alm}
\end{eqnarray}
where $f:~\mathbb{R}^n \mapsto \mathbb{R}$ and $h:~\mathbb{R}^n \mapsto \mathbb{R}^m$ are both convex functions. Solving an unconstrained optimization problem is typically much easier. To this aim, in ALM we instead optimize the following augmented Lagrangian function:
\begin{eqnarray}
\mathcal{L}(X, Y, \mu) = f(X) + \left< Y, h(X) \right> + \frac{\mu}{2} \parallel h(X) \parallel^2_F,
\label{eqn:alm2}
\end{eqnarray}
where $\mu$ is a positive scalar, controlling the strength of original constraints in~(\ref{eqn:alm}). Each iteration optimizes the augmented Lagrangian function and passes the updated $X$, $Y$ values as the initialization of next iteration. The initial value of $\mu$ is exponentially increased until the constraints finally rigidly hold.

%We may apply the augmented Lagrange multiplier method by identifying $X = (A,~E)$ and:
%\begin{eqnarray}
%f(X) = \parallel A \parallel_\ast + \lambda \parallel E \parallel_1,~~h(X) = D - A - E
%\end{eqnarray}
%
%Putting all together, we obtain:
%\begin{eqnarray}
%\mathcal{L}(A,E,Y,\mu) &=& \parallel A \parallel_\ast + \lambda \parallel E \parallel_1 + <Y, D-A-E> \nonumber \\
%&&+ \frac{\mu}{2} \parallel D-A-E \parallel^2_F.
%\end{eqnarray}

%\begin{algorithm}[tb]
%\label{alg:1}
%   \caption{RPCA via the Inexact ALM Method}
%\begin{algorithmic}[1]
%   \STATE {\bfseries Input:} Observation matrix $D \in \mathbb{R}^{m \times n}$, $\lambda$.
%   \STATE $Y_0 = D/J(D)$; $E_0=0$; $\mu_0 > 0$; $\rho > 1$; $k=0$.
%   \WHILE {not converged}
%   \STATE // Line 4-5 solve $A_{k+1} = \arg \min_A \mathcal{L}(A, E_k, Y_k, \mu_k)$.
%   \STATE $(U,S,V) = svd(D - E_k + \mu_k^{-1} Y_k)$;
%   \STATE $A_{k+1} = U S_{\mu_k^{-1}} [S] V^T$.
%   \STATE // Line 7 solves $E_{k+1} = \arg \min_E \mathcal{L}(A_{k+1}, E, Y_k, \mu_k)$.
%   \STATE $E_{k+1} = S_{\lambda \mu_k^{-1}} [D - A_{k+1} + \mu_k^{-1} Y_k].$
%   \STATE $Y_{k+1} = Y_k + \mu_k (D - A_{k+1} - E_{k+1})$;
%   \STATE $\mu_{k+1} = \rho \mu_k$; $k \rightarrow k+1$.
%   \ENDWHILE
%   \STATE {\bfseries Output:} $(A_k, E_k)$.
%\end{algorithmic}
%\label{alg:rpca}
%\end{algorithm}

Unfortunately, directly applying aforementioned technique to~(\ref{eqn:rpca2}) fails to reduce the optimization effort. The resultant augmented Lagrangian function has no closed-form update for variable $A$, mainly due to the matrix nuclear norm $\|A\|_\ast$. Therefore we further relax~(\ref{eqn:rpca2}) by introducing another auxiliary variable $Z$, as follows:
\begin{eqnarray}
\label{eqn:alm3}
\min_{A,E} && \parallel A \parallel_\ast + \lambda \parallel W \otimes E \parallel_1 \\
s.t. && W \otimes D = W \otimes Z + W \otimes E  \\
&& A = Z
\end{eqnarray}

The augmented Lagrangian function for~(\ref{eqn:alm3}) can be accordingly represented as:
\begin{eqnarray}
\label{eqn:alm4}
&&\mathcal{L}(A,E,Y,Z,\mu) \nonumber \\
&=& \parallel A \parallel_\ast + \lambda \parallel W \otimes E \parallel_1 \nonumber \\
&& + <Y_1, W \otimes (D-Z-E)> + <Y_2, A-Z>  \nonumber \\
&& + \frac{\mu}{2} \parallel W \otimes (D-Z-E) \parallel^2_F + \frac{\mu}{2} \parallel A-Z \parallel^2_F,
\end{eqnarray}
which is convex with respect to the variables to be optimized. Here we adopt an alternating minimization strategy.
Each optimization iteration consists of four steps. In each step, a variable is updated in closed form with others fixed.

\textbf{Step-\Rmnum{1}}: Update $A$. The objective function in this step can be described as below:
\begin{eqnarray}
A = \arg \min_A \frac{1}{\mu} \parallel A \parallel_\ast + \frac{1}{2} \parallel A - (Z+Y_2/\mu) \parallel^2_F
\end{eqnarray}

The optimal solution is achieved in closed form base on Singular Value Decomposition (SVD) on matrix $A$. Given a matrix with size $m \times n$, it is well known that full SVD has the complexity of $\mathcal{O}(\min(m n^2), \min(m^2 n))$. See Appendix for details.

\textbf{Step-\Rmnum{2}}: Update $Z$. By setting the first-order derivative of~(\ref{eqn:alm4}) to be zero, it is possible to obtain the following updating rule:
\begin{eqnarray}
%(W^2 + 1 1^T) \otimes Z = W^2 \otimes (D-E) + A + \frac{1}{\mu} (W \otimes Y_1 + Y_2),
W^2 \otimes Z + Z = W^2 \otimes (D-E) + A + \frac{1}{\mu} (W \otimes Y_1 + Y_2),
\end{eqnarray}
where $W^2$ denotes the abbreviation of $W \otimes W$. It is trivially observed that the optimal $Z$ is obtained from direct element-by-element matrix algorithmic operations such as multiplication and division.

%\begin{eqnarray}
%Z = \bar W^{-1} \otimes \Big[ W^2 \otimes (D-E) + A + \frac{1}{\mu} (W \otimes Y_1 + Y_2) \Big]
%\end{eqnarray}
\textbf{Step-\Rmnum{3}}: Update $E$. For clarity, denote $\widetilde E = W \otimes E$, $\widetilde D = W \otimes D$ and $\widetilde Z = W \otimes Z$. It is trivial to obtain the optimal $E^\ast$ from the optimal $\widetilde E^\ast$, therefore we only show the objective function with respect to $\widetilde E$ as below:
\begin{eqnarray}
\widetilde E = \arg \min_{\widetilde E} \frac{\lambda}{\mu} \parallel \widetilde E \parallel_1 + \frac{1}{2} \parallel \widetilde E - (\widetilde D - \widetilde Z + Y_1/\mu) \parallel^2_F.
\end{eqnarray}

The above optimization problem also has closed-form solution with linear complexity. Refer to Appendix for details.

\textbf{Step-\Rmnum{4}}: Update $Y$. The Lagrangian variables are routinely updated as follows:
\begin{eqnarray}
Y_1 &=& Y_1 + \mu(D-Z-E)  \\
Y_2 &=& Y_2 + \mu(A-Z)
\end{eqnarray}

At the end of each iteration, the value of $\mu$ will be increased (\eg, $\mu = \min(\rho \mu, u_{max})$) to tighten the constraints, where $\rho > 1$ is a free parameter. The optimization procedure terminates when the gain of objective function is tiny enough, \ie,
\begin{eqnarray}
\frac{\|A^{k-1}-A^k\|_F+\|E^{k-1}-E^k\|_F}{\|A^{k-1}\|_F+\|E^{k-1}\|_F} \le \epsilon,
\end{eqnarray}
where $A^k, E^k$ denote the estimation of $A,E$ at the $k$-th iteration respectively. $\|\cdot\|_F$ is the Frobenius norm and $\epsilon$ is pre-specified threshold (fixed to be $10^{-4}$ in our implementation).

%stopping conditions: $\parallel D-Z-E \parallel_\infty < \epsilon$ and $\parallel A-Z \parallel_\infty < \epsilon$

\section{Experiments}

\subsection{Dataset Description}
\label{subsec:data}

\begin{figure}[t!]
\centering
\includegraphics[width=0.95\linewidth]{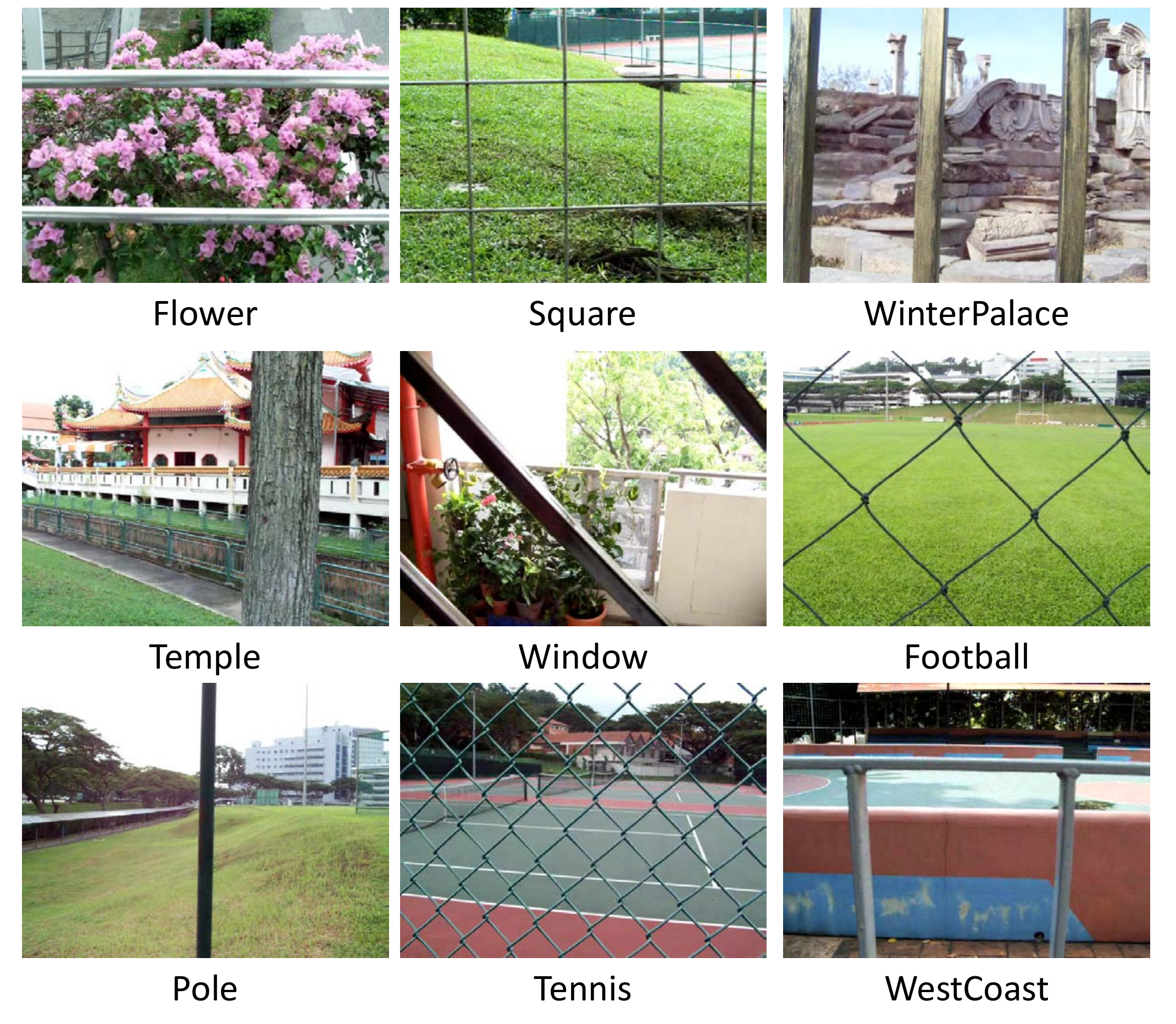}
\caption{\footnotesize The video clips used in our evaluations.}
\label{fig:data}
\end{figure}

To evaluate the proposed algorithm, we capture nine video clips using a Kodak Z650 camera. All of the scenes are in Asia (Beijing or Singapore). Fig.~\ref{fig:data} shows the exemplar frames from these video clips. Most of the video clips have short time duration (from 2 seconds to 8 seconds) and are generally captured following the rules as described in Section~\ref{sec:overview}. We generalize the term of ``fence" to anything that is distant to the target scene. For example, the pole in the video ``Pole" or the tree in the video ``Temple". Obviously it can be hardly epitomized by any kind of image regularities, therefore frustrating the methods in~\cite{liuyanxi08,park10}.

\subsection{Investigation on R-TMF}
\label{subsec:exp-rtmf}

In Fig.~\ref{fig:tmf2} we compare the resultant quality of different estimators on aligned frames, together with the result obtained by the build-in image completion utility in PhotoShop. The superior quality of our proposed method proves the effectiveness of temporal information in video de-fencing, and also highlights the necessity of data weighting towards robust estimation.

%More experimental results are found in Fig.~\ref{fig:exp}.

%\begin{figure}[thb]
%\centering
%\subfigure{
%\includegraphics[width=0.45\linewidth]{./figures/fb25_mask.jpg}
%}
%\subfigure{
%\includegraphics[width=0.45\linewidth]{./figures/fb25_map.jpg}
%}
%\caption{\footnotesize Left: active pixels (in white) in R-TMF stage. Right: unchanged pixels (in white).}
%\label{fig:map}
%\end{figure}
%
%It is computationally expensive to throw all the pixels to the R-TMF. Instead, we threshold by their PoF values. Fig.~\ref{fig:map} shows the mask of the video frame in Fig.~\ref{fig:trunc}. To convey enough contextual information, the set of active pixels should be moderately large. Recall that matrix $A$ resulting from R-TMF is low-rank and tends to be smooth. To enhance the visual plausibility of restored videos, we calculate the luminance difference between the original and restored values. Those pixels with only slight changes will stick to their original values. See the right panel of Fig.~\ref{fig:map}.

\subsection{More Results of Restored Frames}
\label{subsec:restore}

%\begin{figure*}[thb]
%\centering
%\subfigure[Original Frame 15]{
%\includegraphics[width=0.21\linewidth]{./figures/wp_15.jpg}
%}
%\subfigure[Original Frame 45]{
%\includegraphics[width=0.21\linewidth]{./figures/wp_45.jpg}
%}
%\subfigure[Original Frame 2]{
%\includegraphics[width=0.21\linewidth]{./figures/fb_2.jpg}
%}
%\subfigure[Original Frame 25]{
%\includegraphics[width=0.21\linewidth]{./figures/fb_25.jpg}
%}
%\subfigure[Restored Frame 15]{
%\includegraphics[width=0.21\linewidth]{./figures/wp_15_processed.jpg}
%}
%\subfigure[Restored Frame 45]{
%\includegraphics[width=0.21\linewidth]{./figures/wp_45_processed.jpg}
%}
%\subfigure[Restored Frame 2]{
%\includegraphics[width=0.21\linewidth]{./figures/fb_2_processed.jpg}
%}
%\subfigure[Restored Frame 25]{
%\includegraphics[width=0.21\linewidth]{./figures/fb_25_processed.jpg}
%}
%\caption{\footnotesize More video de-fencing results.}
%\label{fig:exp}
%\end{figure*}

\begin{figure*}[t!]
\centering
\includegraphics[width=0.85\linewidth]{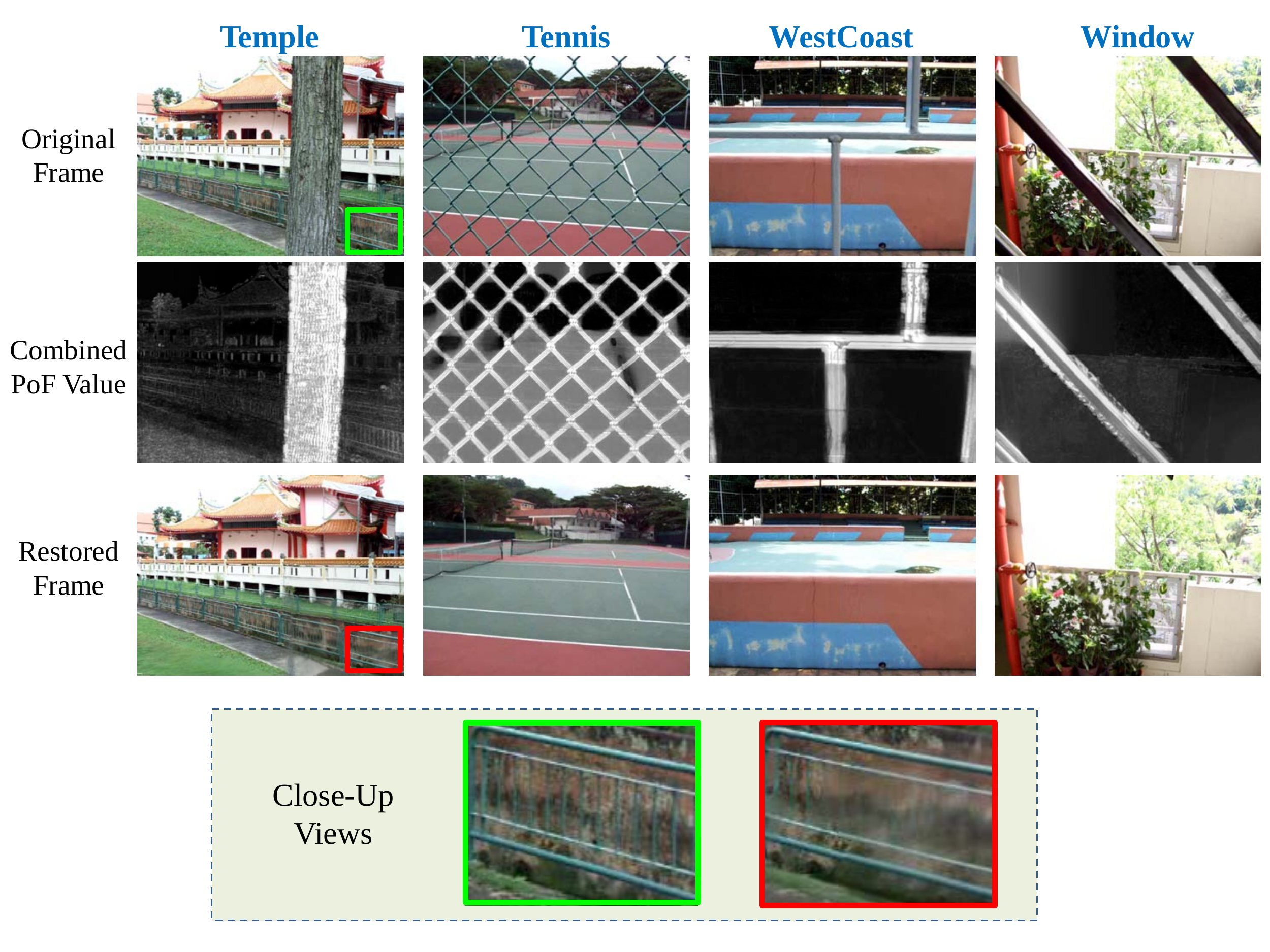}
\caption{\footnotesize More video de-fencing results. See text for more details.}
\label{fig:exp}
\end{figure*}

Fig.~\ref{fig:exp} presents more restored frames for the adopted dataset, where the first three rows display the original video frames, the estimated probability-of-fence values (linear combination of motion cue and appearance cue with equal weights), and the restored frames using the proposed method. The adopted video clips cover a large spectrum of real-world scenes, which demonstrably shows the effectiveness of our proposed framework.

Note that the proposed algorithm has various parameters in different stages, for example, the parameters to estimate the optical flow field and the constant used in Eqn. (\ref{eqn:fa}). Empirically we find that the final results are quite stable over most of the parameters. The only exception is the parameter $M$, which controls the number of consecutive frames used for pixel restoration. As illustrated in Fig.~\ref{fig:camera2}, the optimal value of parameter $M$ is related to several factors, including the spatial extent of the fence-like objects, the motion speed etc. When the ``fence" has a wide span (\eg, the tree in the video clip ``Temple"), the algorithm is still possible to recover the occluded pixels. However, it requires a larger parameter of $M$ (\ie, the number of frames that is required to align to the target frame. See Section~\ref{subsec:alignment}). For the video clip ``Temple", we set $M = 13$ (by default $M = 7$ for others) to obtain the restored results in Fig.~\ref{fig:exp}.

Larger $M$ is a double-edge sword. It enables the recovery of more pixels, and simultaneously complicates accurate alignment of all $2 M + 1$ frames. Fig.~\ref{fig:exp} displays a local image region, where heavy blur is observed. Note that the depth of the enlarged region is very close to the ``fence". Since the image is mainly aligned with respect to the target scene, the mis-alignment for the enlarged region is understandably increased.

\begin{figure}[t!]
\centering
\includegraphics[width=0.75\linewidth]{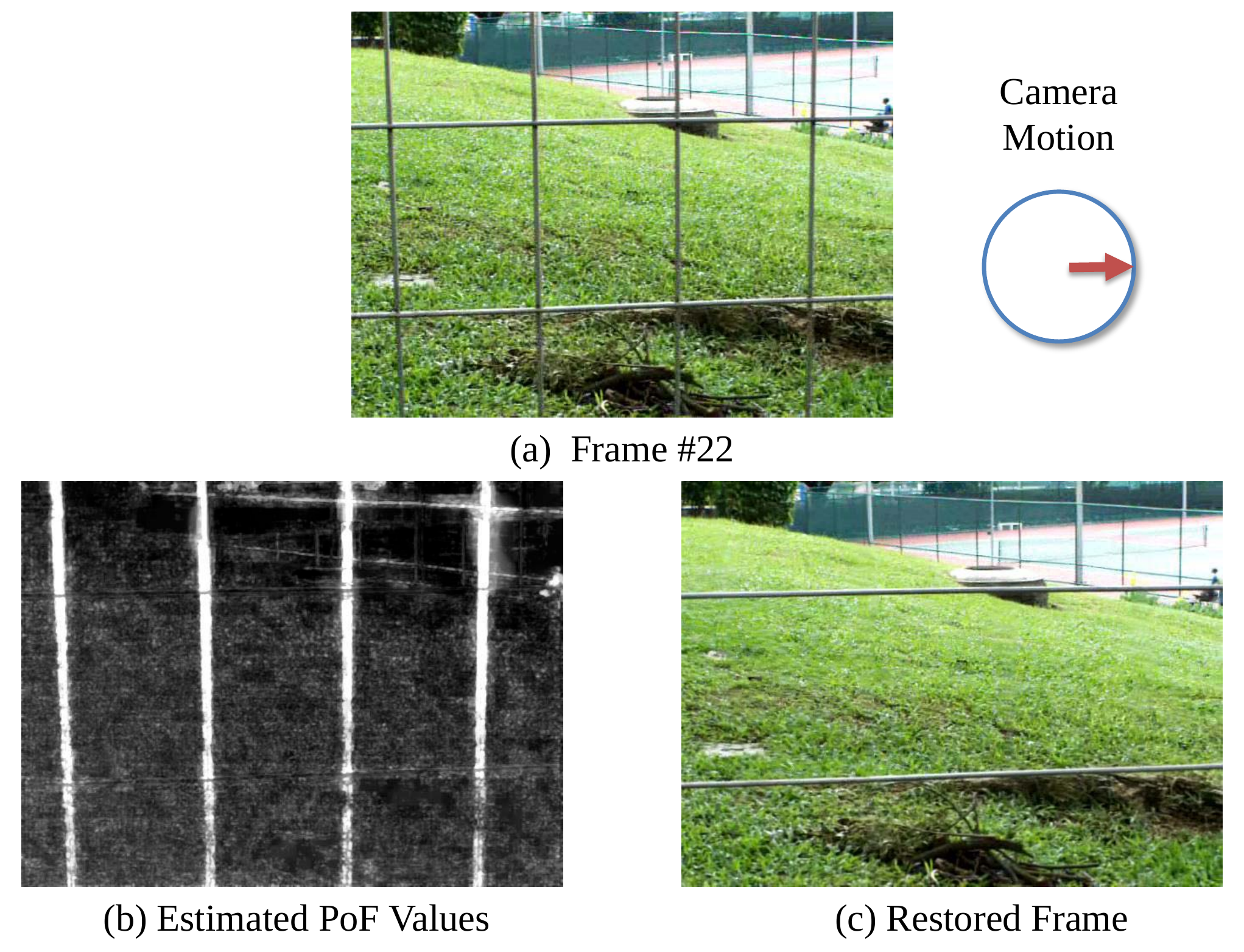}
\caption{\footnotesize A failure case caused by the inconsistency of fence geometry and camera motion. The camera undergos a horizontal move, which will make the horizontal fences impossible to be identified.}
\label{fig:exp2}
\end{figure}

The only failure case among the nine video clips is the one named ``Square". As shown in Fig.~\ref{fig:exp2}, the algorithm fails to recover the horizontal bars. It mainly results from the inconsistency between fence geometry and camera motion, which violates the hypothesis presented in Section~\ref{sec:overview}. However, as shown on other video clips, when those hypothesis are satisfied, the algorithm works reasonably well.

\subsection{Complexity}
\label{subsec:complexity}

Regarding the computational efficacy, the most time-consuming operations are the frame alignment in Section~\ref{subsec:alignment} and R-TMF estimator. Our captured videos all have the resolution of $480 \times 360$ pixels and roughly contain 40 consecutive frames. The optical flow computation between two frames roughly takes 2.2 seconds on our desktop computer equipped with 8G bytes memory and Intel Q9559 CPU. Overall $2 M$ (typically $M=7$) optical flow optimizations are involved during frame alignment, which indicates a rough time cost of 30 seconds. Another $~40$ seconds are spent on R-TMF estimator. Generally the restoration of each single frame is accomplished within 80 seconds.

\section{Conclusions, Limitations and Future Perspective}
\label{sec:conclusion}

In this paper we present a new research topic, the so-called video de-fencing, and propose a framework based on parallax-aware sub-pixel frame alignment and robust pixel restoration. The current solution is focusing on the videos with static scenes. We evaluate the proposed method on a set of real-world consumer videos and generate promising results on most of them. The proposed robust temporal median filter (R-TMF) is a general tool that can be applied in numerous applications.

%Moreover, the proposed method is also potentially useful for raindrop or snowflake removal, since these tasks share the above-mentioned two ingredients with video de-fencing.

The limitation of our work in this paper mainly lies in the incapability of handling moving objects, since a moving object will disrupt the estimation of depth of field from pixel displacement on the video frames. Likewise, the proposed method has difficulty when the background scene and fence-like objects have similar depths. Another down side of the proposed method is the requirement of video capturing as introduced in Section~\ref{sec:overview}.

Regarding the future work, the proposed framework is expected to be extended along the following directions:
\begin{itemize}
\item Extension to the dynamic scenes. They are challenging, since it is hard to distinguish the parallax caused by depth discrepancy or object motion. With high probability the moving objects will be judged as ``fence" and removed. To disambiguate these two kinds of parallax, additional cues will be used. For example, it can be assumed that pixels from the fence are homogeneously subject to specific appearance model (\eg, color-based Gaussian mixture model), which excludes the pixels from moving objects. Another possible solution is manually specifying the mask of the fence at several key-frames.
%\item A user-friendly interface to take human into the loop. Recall that our proposed solution is purely automated. Users can be involved to provide additional cues. For example, the
\item Enhanced sub-pixel image alignment with additional constraints. Conventional optical flow methods seldom target those scenes with diverse depth as in the application of video de-fencing. Additional constraints (\eg, SIFT point correspondence) can be further incorporated to enhance the accuracy of image alignment. Moreover,  it is also helpful to develop a user-friendly interface that takes users into the loop.
%\item background-foreground separation
\item Public benchmark for quantitative comparison. In this work we construct a comprehensive video set for qualitative evaluations. However, it is difficult to obtain the ground truth videos (\ie, videos captured under the same settings yet without fence-like objects) for such real-world videos. In the future we will try to establish a public benchmark from artificially fenced sequences for comparing different algorithms developed for the video de-fencing task.
\end{itemize}

\section*{Appendix}

In this section we introduce two optimization problems mentioned in Section~\ref{subsec:rtmf}. First we define the following soft-thresholding operator:
\begin{eqnarray}
S_\epsilon[x] = \left\{ \begin{array}{ll} x-\epsilon, & \mbox{if}~~ x > \epsilon \\ x + \epsilon,& \mbox{if}~~ x < -\epsilon \\ 0, & \mbox{otherwise} \end{array} \right.
\end{eqnarray}

The effect of $S_\epsilon[x]$ is shrinking $x$ towards zero, controlled by parameter $\epsilon$. Based on the above operator, it is possible to pursue the closed-form solutions for the following optimization problems~\cite{cai10}:
\begin{eqnarray}
U S_\epsilon[S] V^T &=& \arg \min_X \epsilon \|X\|_\ast + \frac{1}{2} \| X - W \|^2_F, \\
S_\epsilon[W] &=& \arg \min_X \epsilon \|X\|_1 + \frac{1}{2} \|X-W\|^2_F,
\end{eqnarray}
where $W = U S V^T$ denotes the SVD of $W$.


\end{document}